\documentclass[runningheads]{llncs}

\usepackage{wrapfig}

\usepackage[mobile]{eccv}

\usepackage{eccvabbrv}

\usepackage{caption}
\usepackage{tablefootnote}

\usepackage{graphicx}
\usepackage{booktabs}
 \usepackage{multirow}
 \usepackage{diagbox}
\usepackage{pifont}
\newcommand{\cmark}{\ding{51}}%
\newcommand{\xmark}{\ding{55}}%

\newcommand{\vc}[1]{\mathbf{#1}}
\usepackage[accsupp]{axessibility}  

\usepackage{hyperref}

\usepackage{orcidlink}

\begin{document}

\title{Pose-Aware Self-Supervised Learning with Viewpoint Trajectory Regularization}

\titlerunning{Pose-Aware SSL}

\author{%
\setlength{\tabcolsep}{10pt}
\begin{tabular}{@{}ccc@{}}
Jiayun Wang$^{1}$
&
Yubei Chen$^{2}$&  
Stella X. Yu$^{1,3}$
\\
 $^1$UC Berkeley&   
 $^2$UC Davis& 
 $^3$University of Michigan\\
\tt\small peterwg@berkeley.edu & 
\tt\small ybchen@ucdavis.edu &
\tt\small stellayu@umich.edu \\
\end{tabular}
}
\authorrunning{Wang et al.}
\institute{}
\maketitle

 \vspace{-0.5em}
\begin{abstract}

Learning visual features from unlabeled images has proven successful for semantic categorization, often by mapping different {\it views} of the same object to the same feature to achieve recognition invariance.  However, visual recognition involves not only identifying {\it what} an object is but also understanding {\it how} it is presented.  For example, seeing a car from the side versus head-on is crucial for deciding whether to stay put or jump out of the way.  While unsupervised feature learning for downstream viewpoint reasoning is important, it remains under-explored, partly due to the lack of a standardized evaluation method and benchmarks.
\\
We introduce a new dataset of adjacent image triplets obtained from a viewpoint trajectory, without any semantic or pose labels.  We benchmark both semantic classification and pose estimation accuracies on the same visual feature.    Additionally, we propose a viewpoint trajectory regularization loss for learning features from unlabeled image triplets.  Our experiments demonstrate that this approach helps develop a visual representation that encodes object identity and organizes objects by their poses, retaining semantic classification accuracy while achieving emergent global pose awareness and better generalization to novel objects.  Our dataset and code are available at \url{http://pwang.pw/trajSSL/}.


\end{abstract}

 \section{Introduction}
\label{sec:intro}

Learning visual features from unlabeled images has proven successful for semantic categorization.  Compared to supervised feature learning, self-supervised learning (SSL) can discover data patterns without labels \cite{wu2018unsupervised,he2020momentum,chen2020simple,bardes2021vicreg},
improve the performance of large-scale vision and language models \cite{achiam2023gpt,bai2024sequential}, 
remain highly flexible \cite{oquab2023dinov2} and generalizable to real-world data \cite{pantazis2021focus,wang2022unsupervised}.

SSL methods so far have focused on coarse-grained recognition, by mapping different {\it views} of the same object to the same feature to achieve recognition invariance \cite{wu2018unsupervised,he2020momentum,chen2020simple,bardes2021vicreg}.  Consequently, both task-specific \cite{wang2022unsupervised} and foundational models \cite{el2024probing} are poor at recognizing objects with unseen or rare poses.  Most data collections do not evenly cover the full range of object poses, while the training data is pivotal for robust performance \cite{schuhmann2022laion,li2024datacomp}.  Lacking pose awareness makes SSL methods worse at generalizing to novel poses.

However, visual recognition involves not only identifying {\it what an object is} but also understanding {\it how it is presented}.  For example, seeing a car from the side versus head-on is crucial for deciding whether to stay put or jump out of the way.  While unsupervised feature learning for downstream viewpoint reasoning is important, SSL is evaluated mostly on semantic tasks, e.g. classification and detection, and its effectiveness for pose-aware representation learning remains under-explored \cite{dangovski2021equivariant}, without a standardized evaluation method.

\begin{figure}[t!]
\centering
\vspace{-1em}
\includegraphics[width=\textwidth]{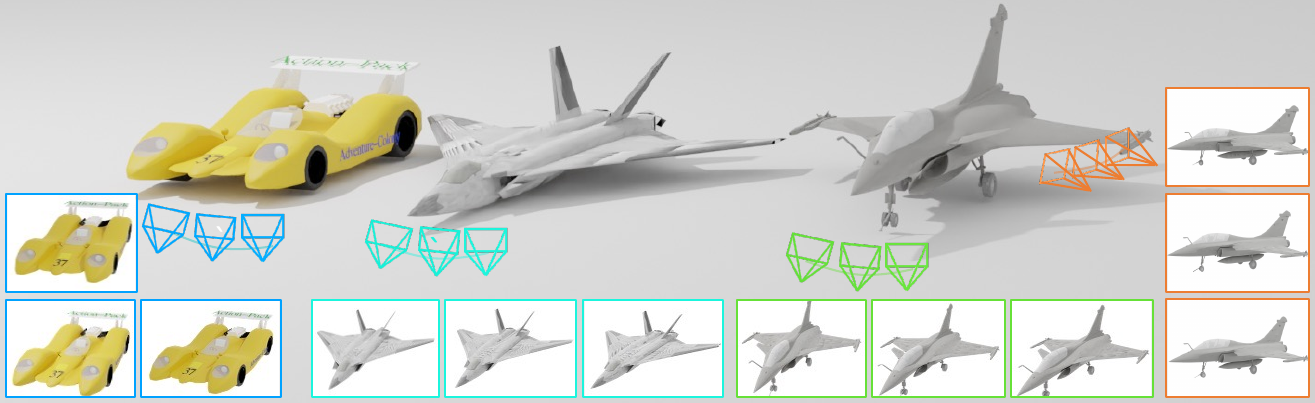}\\
\begin{tabular}{p{\textwidth}}
\setlength{\tabcolsep}{0pt}
{\bf a)} Our training data are {\it unlabeled} image triplets with small pose changes from viewpoint trajectories, without any semantic or pose labels.
\end{tabular}\\[5pt]
\begin{tabular}{@{}p{0.48\textwidth}@{\hspace{12pt}}p{0.48\textwidth}@{}}
\begin{tabular}{@{}l@{}}\setlength{\tabcolsep}{0pt}
\includegraphics[width=0.48\textwidth]{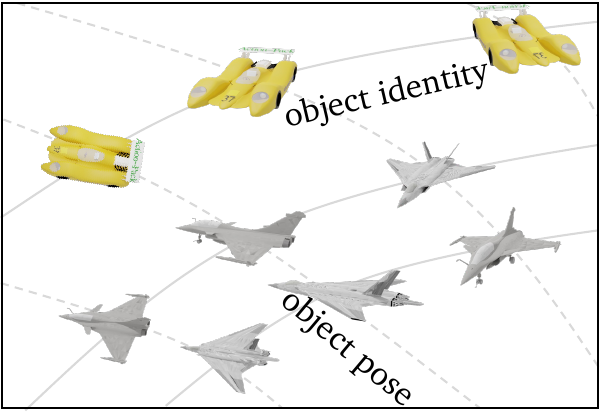}
\end{tabular}&
\begin{tabular}{@{}r@{}}\setlength{\tabcolsep}{0pt}
\includegraphics[width=0.48\textwidth]{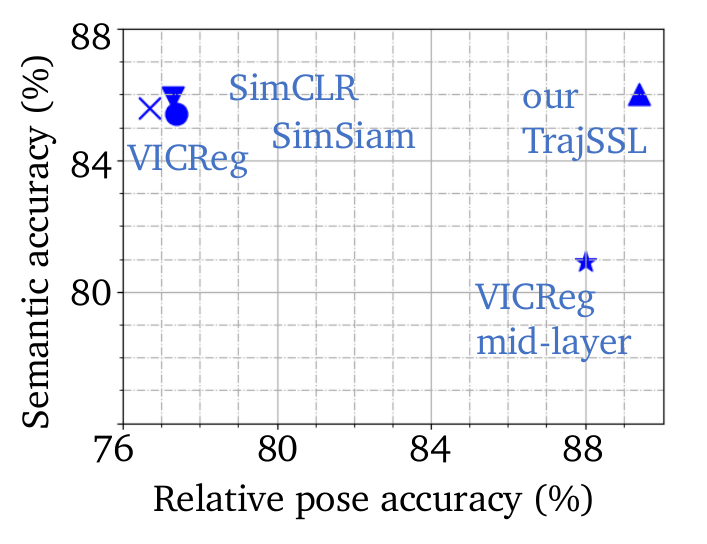}
\end{tabular}\\
{\bf b)} Our learned, emergent representation with ideally disentangled semantics/pose & 
{\bf c)} Our method excels at {\it both} semantic classification and pose estimation.\\
\end{tabular}
\caption{Our goal is to capture two aspects of object recognition through SSL: {\it what the object is} and {\it how the object is presented}.
While the former has been well studied \cite{chen2020simple,bardes2021vicreg}, the latter is rarely understood.
We learn SSL representations that not only capture object semantics but also pose.
{\bf a)} The training data are image triplets with subtle viewpoint changes of objects. The object identity, semantics and pose are unknown.
{\bf b)} The learned representations are expected to discriminate different object semantics and poses, achieving high accuracies for both semantic classification and pose estimation. Notably, we expect to understand {\it global pose} from {\it local pose} changes.
{\bf c)} Our approach improves pose estimation accuracy over existing methods \cite{bardes2021vicreg,chen2020simple,chen2020simsiam} by encouraging images with similar poses to form smooth trajectories in the representation space.
}
\vspace{-1em}
\label{fig:image1}
\end{figure}

We extend the concept of SSL to visual recognition beyond coarse semantic categorization.  We aim at learning a pose-aware visual representation from naturally available visual data, so that it can support down-stream semantic classification and viewpoint estimation (Fig.\ref{fig:image1}).

We first introduce a new object-centric dataset of adjacent image triplets obtained from a viewpoint trajectory, without any semantic or pose labels.  Such a data acquisition scheme is most natural: 1) In human vision, even during fixation at a stationary object, our eyes make continuous and minute movements, including tremor, drift, and microsaccade; 2) In robotic vision, as the robot moves around in the environment, it captures adjacent images of the same object from a smooth viewpoint trajectory.  We generate synthetic image triplets of various objects with slight camera pose changes.

We then benchmark semantic classification and pose estimation on the feature learned from unlabeled image triplets, for seen and unseen objects, as desirable for video analysis \cite{grauman2022ego4d}, robotics \cite{du2021vision} and world-models \cite{sergeant2023influence}.  
Our benchmark precludes the use of semantic or pose labels during training and encompasses both semantic classification and pose estimation tasks during evaluation, unlike existing settings \cite{garrido2023self,lee2021improving} which allow training SSL with pose labels.  

We include both absolute and relative pose estimation tasks.
The former is useful for testing how well SSL learns a global pose from adjacent poses, whereas the latter is useful for testing how well SSL generalizes to out-of-domain poses.   Without defining category-specific canonical poses, SSL can be flexibly evaluated on out-of-domain data and unseen semantic categories.

We benchmark existing SSL methods with a ResNet backbone.  We discover that intermediate-layer features outperform later-layers by absolute 10-20\% gains in pose estimation, at reduced semantic classification accuracies.  This result is not surprising, as the last-layer feature under the SSL objective becomes more invariant to the object pose and
tuned to the semantic category.

We further improve the performance by proposing a viewpoint trajectory regularization loss on intermediate features.
Inspired by slow feature analysis \cite{foldiak1991learning, wiskott2002slow, goroshin2015learning, chen2018sparse},
we encourage adjacent views in the triplet to form a smooth trajectory in the feature space, implemented with a simple local linearity assumption.  

Our simple approach leads to an additional 4\% gain in pose estimation without affecting semantic classification. It is also more effective at out-of-domain generalization and on a real-world rotating-car benchmark Carvana \cite{shaler2017carvana}.

Our work has three main contributions.
{\bf 1)} We introduce a new dataset of unlabeled image triplets and a new SSL benchmark for both semantic classification and pose estimation.
{\bf 2)} We propose a novel viewpoint trajectory regularization loss on intermediate features.  
{\bf 3)} We demonstrate that our simple approach helps develop a visual representation that encodes object identity and organizes objects by pose, retaining semantic classification accuracy while achieving emergent global pose awareness and better generalization to novel objects. 


\vspace{-0.5em}
\section{Related Works}
\vspace{-0.5em}
\label{sec:related}


\textbf{Self-Supervised Learning for Semantic Downstream Tasks}. 
There are predominantly two SSL approaches: contrastive and non-contrastive. Contrastive methods, grounded in the InfoNCE criterion~\cite{oord2018representation}, include  \cite{chen2020simple,he2020momentum,chen2020improved,chen2021mocov3}. A notable variant is clustering-based contrastive learning~\cite{caron2018clustering,caron2021dino,oquab2023dinov2}, which shifts focus from individual samples to cluster centroids. Non-contrastive approaches~\cite{grill2020byol,chen2020simsiam,bardes2021vicreg,ermolov2021whitening,bardes2022vicregl}, on the other hand, aim to align embeddings of positive pairs, similar to contrastive learning, but with strategies to prevent representational collapse.
 Yet, they primarily focus on semantic tasks like semantic classification, leaving geometric tasks such as pose estimation underexplored.
 We bridge the gap by also providing the benchmark for geometric downstream tasks. 
We refer to works above as {\it invariant SSLs} as they learn representations invariant to object pose.

\noindent \textbf{Geometry-Aware Self-Supervised Learning}. 
In the quest for geometry-aware SSL, a prevalent method is to learn equivariant representations.
Past research has utilized autoencoders, including transforming autoencoders~\cite{hinton2011transforming}, Homeomorphic VAEs~\cite{falorsi2018explorations}, or \cite{winter2022unsupervised}.
Recently, EquiMod~\cite{devillers2022equimod} and SEN~\cite{park2022learning} have introduced predictors that enable reconstruction-free representation manipulation in the latent space. 
Another novel approach is learning equivariant representations without prior knowledge of transformation groups, as explored in \cite{shakerinava2022structuring}. 

The most relevant work to ours is SIE \cite{garrido2023self}, which evaluates equivariant representation learning through rotation matrix prediction. Unlike SIE, which uses ground-truth pose labels during training, our approach avoids any geometric or semantic labels in SSL training. Additionally, we assess pose estimation performance on out-of-domain data using relative pose. A comparison of methods is summarized in Table \ref{tab:method_comp} in the supplementary.

\noindent \textbf{Pose Estimation.} 
We adopt pose estimation as a task to evaluate geometric representations, given its fundamental role in many geometry-aware recognition tasks \cite{kappler2018real,macklin2022warp}.
The object pose remains ambiguous unless a canonical pose is defined. However, defining the canonical pose can be challenging for a set of objects with different semantic classes. It is also hard to define a general canonical pose for all categories due to the difficulty of aligning two classes (e.g. airplanes and boats). Relative pose estimation can be used to eliminate the need for canonical pose. 
Specifically, there are two pose estimation evaluation methods:

\noindent{\bf 1) Absolute pose estimation from a single image} is only well-defined if a canonical pose (or canonical coordinate system) exists.  Previous work on single-view pose estimation is therefore class-specific. For a fixed set of categories, they define canonical coordinate systems class-by-class with a prior \cite{kendall2015posenet,kehl2017ssd,iwase2021repose,chen2022occlusion} or learned features \cite{sun2021canonical}. 
In contrast, we propose a class-agnostic approach using $k$ nearest neighbor retrieval ($k$-NN).  Our model identifies the $k$ most similar representations, assuming they belong to the same semantic category. Since all instances within a category adhere to a consistent canonical coordinate system, the predicted pose label derived from $k$-NN will align accordingly.

\noindent {\bf 2) Relative pose estimation from a pair of images} avoids class-specific canonical coordinate system by assuming the first image defines a canonical pose, and thus predicting the relative pose of the second image compared to the first image is not ambiguous. RelPose and its variants \cite{zhang2022relpose,lin2023relpose++}  describe a data-driven method for inferring the relative pose given an image pair, and we adopt their setting for our class-agnostic relative pose estimation.


\section{A Benchmark for SSL Geometric Representations}
\begin{figure}[t!]
  \vspace{-0.5em}
    \centering
    \includegraphics[width=0.99\linewidth]{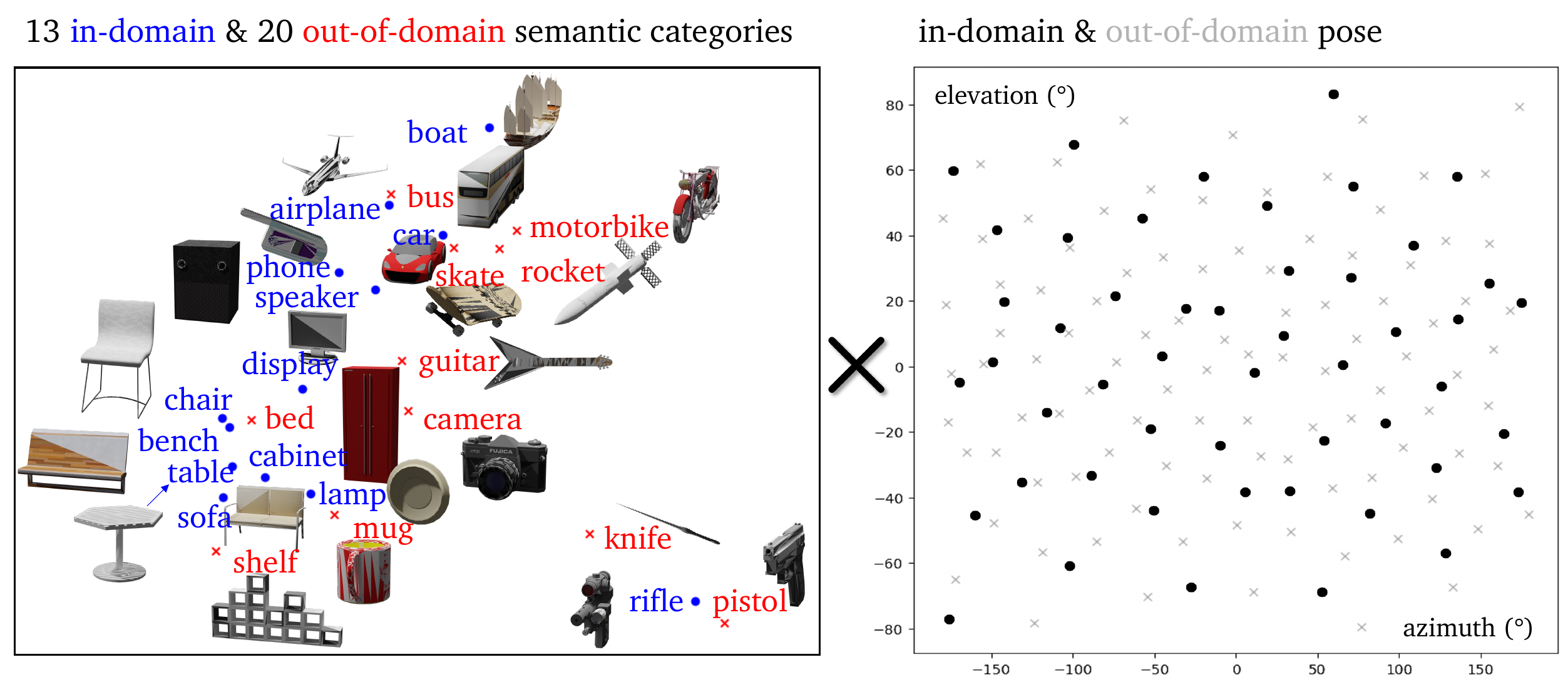}
      \vspace{-1em}
    \caption{
    Our benchmark dataset contains rendered images from ShapeNet \cite{chang2015shapenet}. {\bf Left:}
    For semantics, we use non-overlapping 13 \textcolor{blue}{in-domain}  semantic categories and 11 \textcolor{red}{out-of-domain} categories. We project \textcolor{blue}{in-domain} and \textcolor{red}{out-of-domain} semantic classes with PCA-projected Word2Vec \cite{church2017word2vec} and show a representative object with 
     $(15^{\circ},15^{\circ})$.
    {\bf Right:} For pose, we adopt {\it absolute} and {\it relative} pose estimation as tasks. Notably, relative pose enables SSL's generalizability test on out-of-domain data as it eliminates the need for category-specific canonical pose. The (camera) pose is defined as the spherical coordinate (azimuth, elevation) of the camera position. We render objects from $n$ unique camera angles, uniformly distributed across the viewing sphere $S^2$, utilizing a Fibonacci sphere distribution \cite{alexa2022super}, denoted as $\text{Fib}(n)$ (more details in Fig.\ref{fig:relpose} in supplementary).
    We use $\text{Fib}(50)$ as \textcolor{black}{in-domain} training and \textcolor{gray}{$\text{Fib}(100)$} for \textcolor{gray}{out-of-domain} evaluations.
    In-domain and out-of-domain set statistics are in Table \ref{tab:dataset} in supplementary.} 
    \label{fig:data_vis}
            \vspace{-1.5em}
\end{figure}

We introduce the SSL benchmark with the problem and data setting, as well as the evaluation metrics  (Fig.\ref{fig:data_vis}).

\subsection{The Problem Setting}

We propose a benchmark that evaluates the SSL semantic and geometric representation quality. 
The SSL operates without ground-truth semantic or pose labels during training, aiming to develop representations that are aware of both the semantics and geometry of the input image. Key principles for the SSL benchmark include:
{\bf 1) Training Phase:} SSL is trained purely on images, without any semantic or pose labels. This ensures that all learned information is derived directly from the image itself. 
{\bf 2) Evaluation Phase:} SSL should learn representations that encode both semantic and geometric information. Using different representations from the model for different tasks is fine.

Given the nature of SSL, pose labels are explicitly excluded during training to align with the principle of learning without labels. 
One may argue that in-plane image rotations (as suggested by \cite{dangovski2021equivariant}) could offer pseudo ``pose labels'', but more complex manipulations like 3D rotations are generally unfeasible. We thus strictly avoid any labels during training.
In the evaluation phase, SSL must learn both semantic and geometric representations from data, since both elements and their interplay are essential for a comprehensive understanding of the data and could benefit the overall learning process.
One challenge lies in estimating the pose of an out-of-domain image, as the canonical pose is not defined. We thus introduce relative pose estimation as the metric, with details as follows.







\subsection{Data and Evaluation Metrics}
\label{sec:bench_data}


Our benchmark provides data generation, downstream tasks and evaluation configurations, to evaluate SSL's capacity to capture geometric and semantic information.
For data generation and configuration methods, without loss of generalizability, we use 3D meshes from ShapeNet \cite{chang2015shapenet} to generate images of varied objects in diverse 3D poses as the dataset used empirically in this work.
For the geometric task, we adopt a fundamental one, pose estimation of object-centric images.
We consider both {\it absolute} and {\it relative} 3D pose estimation tasks to enable evaluations on in-domain and out-of-domain data, where we also provide a dataset-splitting configuration.
Compared to existing datasets with similar purposes \cite{zimmermann2021contrastive,garrido2023self}, ours enables
out-of-domain evaluation and the generation method leads to a complete and even pose coverage. Shadows are also not rendered to avoid unintended ground-truth pose information leakage.
We provide a detailed comparison with such datasets in supplementary.

\noindent {\bf Pose}. 
We make source 3D objects of the same semantic class aligned and fixed for image rendering. The pose is defined as the camera pose.
Specifically, we make cameras all reside on a unit $S^2$ sphere  with rendering configurations of
look-at view transform with up vector $(0,1,0)$ and translation vector $(0,0,1)$, following PyTorch3D's convention \cite{ravi2020pytorch3d}, Fig.\ref{fig:data_vis}. 
Camera poses are represented as 
(azimuth, elevation) pairs, the spherical coordinates of camera positions. 
We define the category-specific canonical pose to ensure no {\it absolute pose} ambiguity for in-domain data as objects of each semantic category are aligned. 

\noindent  {\bf Relative Pose} eliminates the necessity of canonical pose by considering two views of an object with  pose $\vc{p_1}, \vc{p_2}$, and is defined as 
 $\vc{\Delta p} =\vc{p_2}- \vc{p_1}$, the pose difference from view 2 to view 1. Introducing relative pose ensures the SSL generalizability evaluation on out-of-domain images where canonical poses are not tractable. This differs from the previous setting \cite{garrido2023self} where out-of-domain data cannot be considered with only absolute pose estimation.


\noindent {\bf Pose Sampling.} 
For uniform camera coverage of the whole viewing sphere, for each object, we use Fibonacci lattices \cite{hardin2016comparison,alexa2022super}, placing $n$ cameras at each lattice point to render $n$ views, denoted as Fib$(n)$. 
We render in-domain poses using Fib$(50)$ and out-of-domain poses with Fib$(100)$, rotating Fib$(100)$ to avoid pose overlap.
 Fig.\ref{fig:data_vis}{\bf Right} depicts rendered images with fixed pose.

\noindent {\bf Dataset Split}. We divide the dataset for in-domain and out-of-domain parts (statistics in Table \ref{tab:dataset} in supplementary). For the pose, we use non-overlapping Fib$(50)$ and Fib$(100)$ for in-domain and out-of-domain poses as mentioned before. For semantic categories, we use 13 object classes (e.g. airplane, car, watercraft, etc.) as {in-domain} data, and  11 object classes (e.g. bed, guitar, rocket, etc.) as {out-of-domain} data.
There is no overlapping for the two sets. For each semantic category, we render 400 different objects, with 320 for unsupervised training (or probe training) and 80 for testing. For simplicity, our experiments focus on cases where either pose or semantics are out-of-domain, not both. 



\noindent{\bf Downstream Tasks and Evaluation}. We follow previous benchmarks for semantic classification.
To evaluate geometric representation, our benchmark includes the following downstream task configurations with ShapeNet \cite{chang2015shapenet} as an example (Fig.\ref{fig:data_vis}):
\noindent {\it 1) In-Domain: Absolute Pose}. Utilizing 13 in-domain semantic categories and poses from Fib$(50)$, we assess absolute pose through nearest neighbor retrieval. 
\noindent {\it 2) In-Domain: Relative Pose}.
This task maintains the same data setting as the in-domain absolute pose, while the distinction lies in the training method. After unsupervised training, we employ a simple probe to train on 80\% of the instances' frozen representations for relative pose estimation, with the remaining 20\% used for performance evaluation.
\noindent {\it 3) Out-of-Domain: Unseen Poses}. 
We work with the 13 in-domain semantic categories but with unseen poses from Fib$(100)$. We only evaluate relative pose estimation performance with a simple probe for faster inference speed.
\noindent {\it 4) Out-of-Domain: Unseen Semantic Categories}. 
This scenario involves 11 unseen semantic categories paired with in-domain poses from Fib$(50)$. Similarly, we evaluate relative pose estimation performance.

\begin{figure}[t!]
        \vspace{-1.3em}
    \centering
    \includegraphics[width=0.96\linewidth]{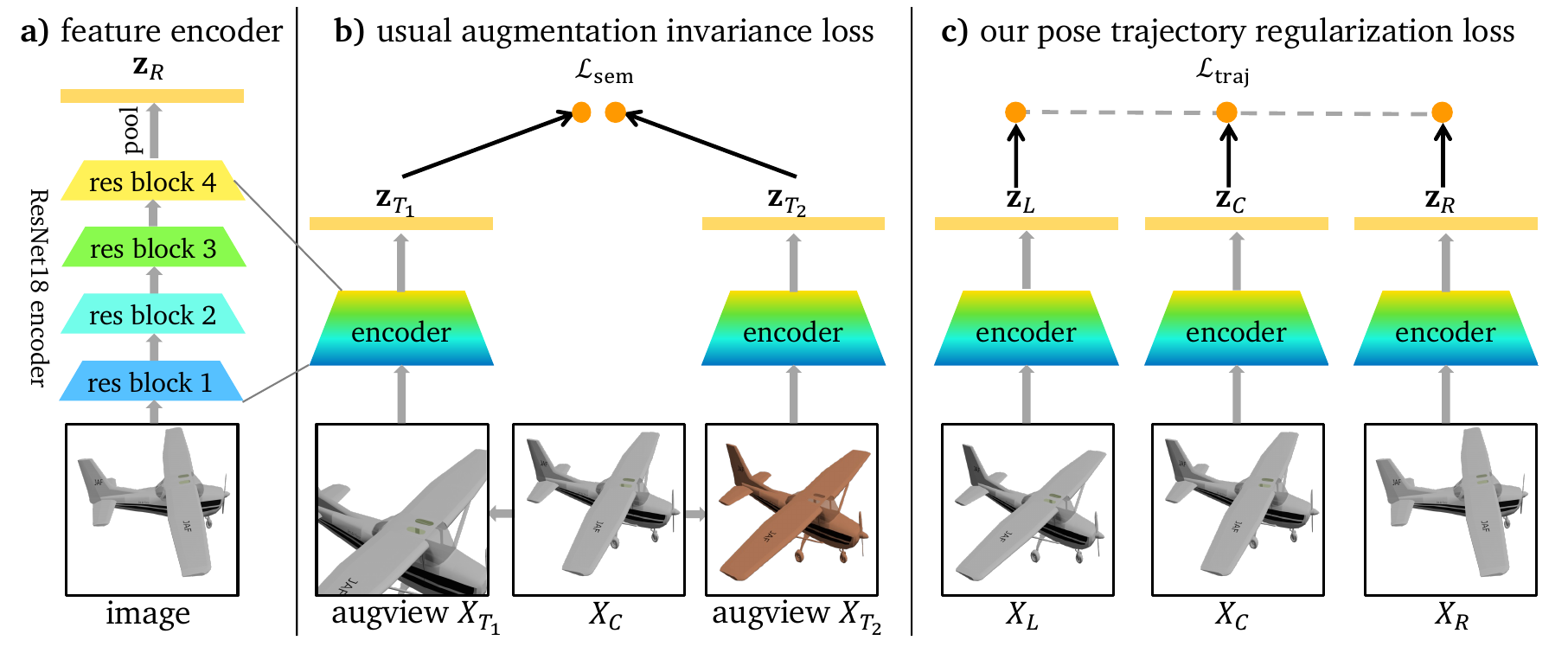}
            \vspace{-1.3em}
 \caption{We impose an unsupervised loss on the feature representations, after feeding the image through an encoder ({\bf a}).
 In addition to an unsupervised semantic loss $\mathcal{L}_{\text{sem}}$ ({\bf b}) which is commonly used in SSL, we add a trajectory loss $\mathcal{L}_{\text{traj}}$ (Eqn.\ref{eq:traj}) ({\bf c}) to enhance geometric representation.
        $\mathcal{L}_{\text{sem}}$ always follows baseline settings, which is applied post-projector for SimCLR \cite{chen2020simple}, for example. $\mathcal{L}_{\text{traj}}$ always operates on the pooled feature $z$. For pose evaluation, we allow representations from different layers and find that mid-layer representations like ``res block3'' give pose estimation gain.}            \vspace{-2em}
                \label{fig:architect}
\end{figure}

\section{Enhancing Geometric Representation Learning}

To enhance geometric representation learning, we first investigate whether earlier layers, rather than the commonly used last layer, are better suited for this task. We then introduce trajectory regularization for image triplets with natural, continuous view changes to further refine geometric representation.

\subsection{Mid-Layer Representation for Evaluation}

We explore the feasibility of using different layers of the backbone to predict pose.
This consideration stems from the understanding that geometric tasks are typically mid-level vision tasks, whereas semantic tasks align with high-level ones. 
Additionally, unlike whole-image embedding which is approximately the average of local patch embeddings \cite{chen2022bag}, mid-level features are local embeddings that could capture mid-level visual cues like pose. Mid-level features can thus be considered as a combination of patch embeddings that enhance pose estimation. 
We focus on whether using representations from mid-layers, such as the ``res block3'' or ``res block4'' layers (referenced in Fig.\ref{fig:architect}), enhances pose estimation performance. For simplicity, we refer to ``res block3'' as ``conv3'' hereafter.

Empirically, our results show a significant improvement in pose estimation, with gains ranging from 10\%-20\% when using mid-layer representations (detailed in Section \ref{sec:midlayer}). 
Further, as a verification of the similarity between mid-layer representations and patch embeddings, we concatenate embeddings of local image patches and observe a similar performance to ``conv4'' embedding ($87\%$ vs $88\%$, details in supplementary).
A common challenge with mid-layer representations is their high dimensionality, primarily due to large spatial sizes. This high dimensionality can lead to inefficiencies during inference and storage. We demonstrate in Section \ref{sec:compression} that high-dimensional mid-layer representations can be effectively compressed with minimal accuracy drop, thereby enhancing overall efficiency.

\subsection{Trajectory Regularization}

Given an image $X$ of an object with pose $\vc{p}$, we feed it to an encoder $f_{\theta}$ to obtain a representation $\vc{z}=f_{\theta}(X)$, which is used for both semantic and geometric tasks.

\noindent {\bf Invariant SSL} refer to methods \cite{chen2020simple,he2020momentum,bardes2021vicreg} that generate representations that are invariant to data augmentations (e.g., random crops), which sometimes include geometric augmentations, due to the primary focus on semantic representations.
For an image $X$, invariant SSLs  create two augmented variants,  $X_{T_1}$ and  $X_{T_2}$, which are then fed into the encoder $f_{\theta}$ for two respective representations, $\vc{z}_{T_1}$ and $\vc{z}_{T_2}$.
The invariant loss, or unsupervised semantic loss is method-dependent and can be denoted as $\mathcal{L}_{\text{sem}}(\vc{z}_{T_1}, \vc{z}_{T_2})$. 
Despite their focus on semantic information, we evaluate such invariant SSL representations for predicting pose $\vc{\hat{p}}$ of image $X$, considering that pose information might be encoded within $\vc{z}$.

\begin{figure}[b]
    \begin{minipage}{0.41\textwidth}
        \centering
        \includegraphics[width=\linewidth]{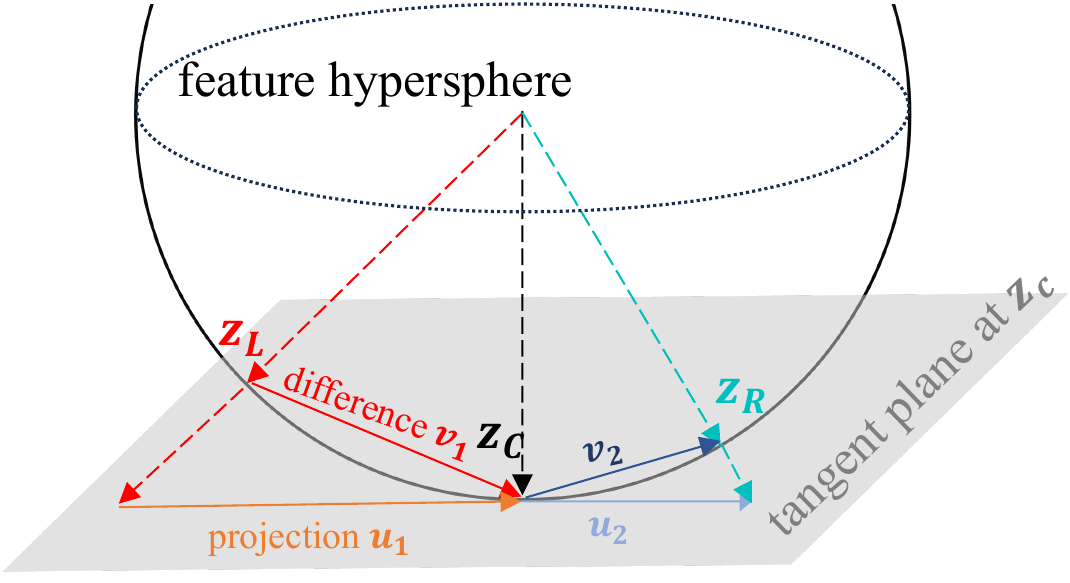} 
    \end{minipage}%
    \hfill
    \begin{minipage}{0.6\textwidth}
        \caption{We enforce representations of adjacent views of an object, $\vc{z_L},\vc{z_C},\vc{z_R}$, to form a geodesic trajectory. {\bf Upper:} $\vc{z}$ resides on a unit hypersphere. The objective is to map the difference vectors $\vc{v_1} = \vc{z_C} - \vc{z_L}$ and $\vc{v_2} = \vc{z_R} - \vc{z_C}$ onto $\vc{z_C}$'s tangent plane, optimizing for maximal cosine similarity to achieve a linear trajectory on that plane. {\bf Lower:} Projected vector $\vc{u}$ is computed by deducting the normal component $\vc{z_C}$ from the  difference vector $\vc{v}$.}
        \vspace{-1.3em}
                \label{fig:linear}
    \end{minipage}
\end{figure}

\noindent {\bf Trajectory-Regularized SSL}.
We aim to enhance the SSL geometric representation quality by leveraging a natural prior:  representations of objects with incremental pose changes should form a smooth, low-curvature path in the representation space. This leads us to promote a locally linear trajectory for representations corresponding to slight pose variations. Linear trajectory requires small camera pose changes only and does not violate the SSL setting. 

Consider a triplet of images $\{X_L, X_C, X_R\}$ from a sequence with  respective poses ${{p_L}, {p_C}, {p_R}}$ forming a trajectory, where pose changes are subtle. That is, $\{X_L, X_C, X_R\}$ form an adjacent pose triplet. The encoded representations $\vc{z}_L, \vc{z}_C, \vc{z}_R$ are normalized and residing on a unit hypersphere (i.e., $|\vc{z}|_2=1$). Our goal is to align these points along a geodesic trajectory on the hypersphere. This is achieved by projecting the difference vectors between representations onto the tangent plane at $\vc{z_C}$, thereby enforcing a linear trajectory (Fig.\ref{fig:linear}).


The difference of two representations with adjacent poses is $\vc{v}_1=\vc{z}_C-\vc{z}_L, \vc{v}_2=\vc{z}_L-\vc{z}_C$. These vectors are projected on the tangent space at $\vc{z}_C$:
\begin{align}
\vspace{-0.5em}
    \mathbf{u}_i &= \mathbf{v}_i - (\mathbf{v}_i \cdot \mathbf{z}_C)\mathbf{z}_C, \qquad i=1, 2 
\vspace{-0.5em}
\end{align}
We then maximize the cosine similarity between $\vc{u}_1$ and $\vc{u}_2$ to enforce linearity in the trajectory. The trajectory loss, or pose loss, is defined as:
\begin{align}
\label{eq:traj}
        \mathcal{L}_{\text{traj}}(\vc{z}_L, \vc{z}_C, \vc{z}_R) = -\frac{\vc{u}_1 \cdot \vc{u}_2}{ \| \vc{u}_1 \| \|\vc{u}_2 \| }
\end{align}

Semantic loss $\mathcal{L}_{\text{sem}}$ is also incorporated for semantic representation capacity. We apply augmentations to $X_C$ to generate $X_{T_1}$ and $X_{T_2}$, then apply semantic loss on their representations $\mathcal{L}_{\text{sem}}(\vc{z}_{T_1}, \vc{z}_{T_2})$. Our total loss combines both semantic and pose losses (Fig.\ref{fig:architect}):
\begin{align} 
\label{totalloss}
    \mathcal{L} = \mathcal{L}_{\text{sem}}(\vc{z}_{T_1}, \vc{z}_{T_2})+\lambda \mathcal{L}_{\text{traj}}(\vc{z}_L, \vc{z}_C, \vc{z}_R)
\end{align}
where weight $\lambda$ balances the trajectory loss.
We always apply the trajectory loss $\mathcal{L}_{\text{traj}}$ (as per Eqn.\ref{eq:traj}) at the pooled feature layer $z$, as empirically altering the layer in $\mathcal{L}_{\text{traj}}$ impacts downstream task performance by only about 1\%.

\section{Experiments}
\label{sec:exp}
\vspace{-0.5em}
We first discuss the training and evaluation protocols, and then report and compare evaluation performance when using the last and mid-feature layer as the representation. We conclude the section with representation visualizations.
\subsection{Training Protocols}
\vspace{-0.5em}

Our experimental framework adopts the common two-stage approach used in SSL. Supervised baselines are included as references. The first stage is unsupervised pretraining (or supervised training), where the model is fed with training data. The specific training protocols of fully-supervised, geometry-supervised and self-supervised methods are method-dependent. The second stage is evaluation, which includes directly evaluating the learned representation on downstream tasks (with the nearest neighbor) and simple probes trained on frozen representations for downstream tasks. The second evaluation stage is the same for all methods for fairness (details in Section \ref{sec:eval_p}). 
We mainly consider three baselines: fully-supervised, geometry-supervised, invariant SSL. 
We discuss the training settings of each method below. 

\noindent {\bf Fully-Supervised Learning.} We provide supervised baselines to establish an upper bound for in-domain performance (Table \ref{tab:feats} in supplementary, first row). Separate models for semantic classification and pose estimation are trained with corresponding ground-truth labels to prevent task interference.

\noindent {\bf Geometry-Supervised Learning.} 
Following previous methods \cite{lee2021improving,dangovski2021equivariant,garrido2023self}, 
baselines are trained on ground-truth pose labels but not semantic labels during training (Table \ref{tab:feats} in supplementary, second row). Specifically, we replicate the AugSelf \cite{lee2021improving} setting, combining an unsupervised semantic loss $\mathcal{L}_{\text{sem}}$ with a cross-entropy loss for pose label classification. The results are for reference and not a direct or fair comparison to SSLs.


\noindent {\bf Invariant Self-Supervised Learning.} We consider two state-of-the-art SSL methods, VICReg \cite{bardes2021vicreg} and SimCLR \cite{chen2020simple}. Images of the same object with different poses are treated as distinct samples. We follow their training settings and use standard data augmentation (e.g. random crop and color jittering).

\noindent {\bf Trajectory-Regularized Self-Supervised Learning.} We add the  trajectory loss  $\mathcal{L}_{\text{traj}}$ to the invariant SSL \cite{chen2020simple,bardes2021vicreg}.
As mentioned earlier, we assume image triplets from a sequence with small relative pose changes are available. 
We implement as follows: during training, for an image $X_C$ with pose $\vc{p_C}$, we randomly select an adjacent left image $X_L$ with pose $\vc{p_L}$. Using slerp \cite{shoemake1985animating}, we obtain the right pose $p_R$ such that $\vc{p_R} - \vc{p_C} = \vc{p_C} - \vc{p_L}$, and render the right image $X_R$. The image triplet $\{X_L, X_C, X_R\}$ can now be used to obtain the trajectory loss. Importantly, no additional transformations like random cropping are applied to ${X_L, X_C, X_R}$ to preserve geometric information.
Our method still works for non-equidistant poses, i.e. $\vc{p_R} - \vc{p_C} \neq \vc{p_C} - \vc{p_L}$, with details in supplementary.

\noindent {\bf Shared Protocols.}  All methods utilize a ResNet-18 \cite{he2016deep} as the backbone encoder\footnote{Our method also works with other model architecture (Section \ref{sec:ablation} in supplementary)}. Training is consistent across models, spanning 300 epochs using the LARS optimizer \cite{you2017large} with a learning rate of 0.3 and weight decay of $10^{-4}$.

\subsection{Evaluation Protocols}
\label{sec:eval_p}

{\bf Semantic Classification}. We evaluate with a linear classification on top of the frozen representation from the feature layer with dimension $512$. 

\noindent {\bf Pose Estimation.}
As a comprehensive evaluation, we consider both absolute and relative pose estimation tasks: 
{\bf 1) Absolute pose estimation.} We employ a weighted $k$-nearest neighbor classifier as used in \cite{wu2018unsupervised} on the representations  from the feature layer. 
{\bf 2) Relative pose estimation.} 
We obtain feature-layer representations $z_1, z_2$ from two different views of an object. These representations are concatenated (resulting in a $1024$-dim feature), and a simple probe of a two-layer perceptron with a hidden dimension of $1024$ is used to predict the relative pose. Relative pose estimation is more computationally efficient but is generally harder as it relies on only two views for inference. 
We consider in-domain and out-of-domain scenarios for pose estimation and only report relative pose performance to avoid redundancy (as mentioned in Section \ref{sec:bench_data}). 



\subsection{Evaluation on Last Feature-Layer}

  \begin{figure}[t!]
  \vspace{-0.5em}
    \centering
    \includegraphics[width=\linewidth]{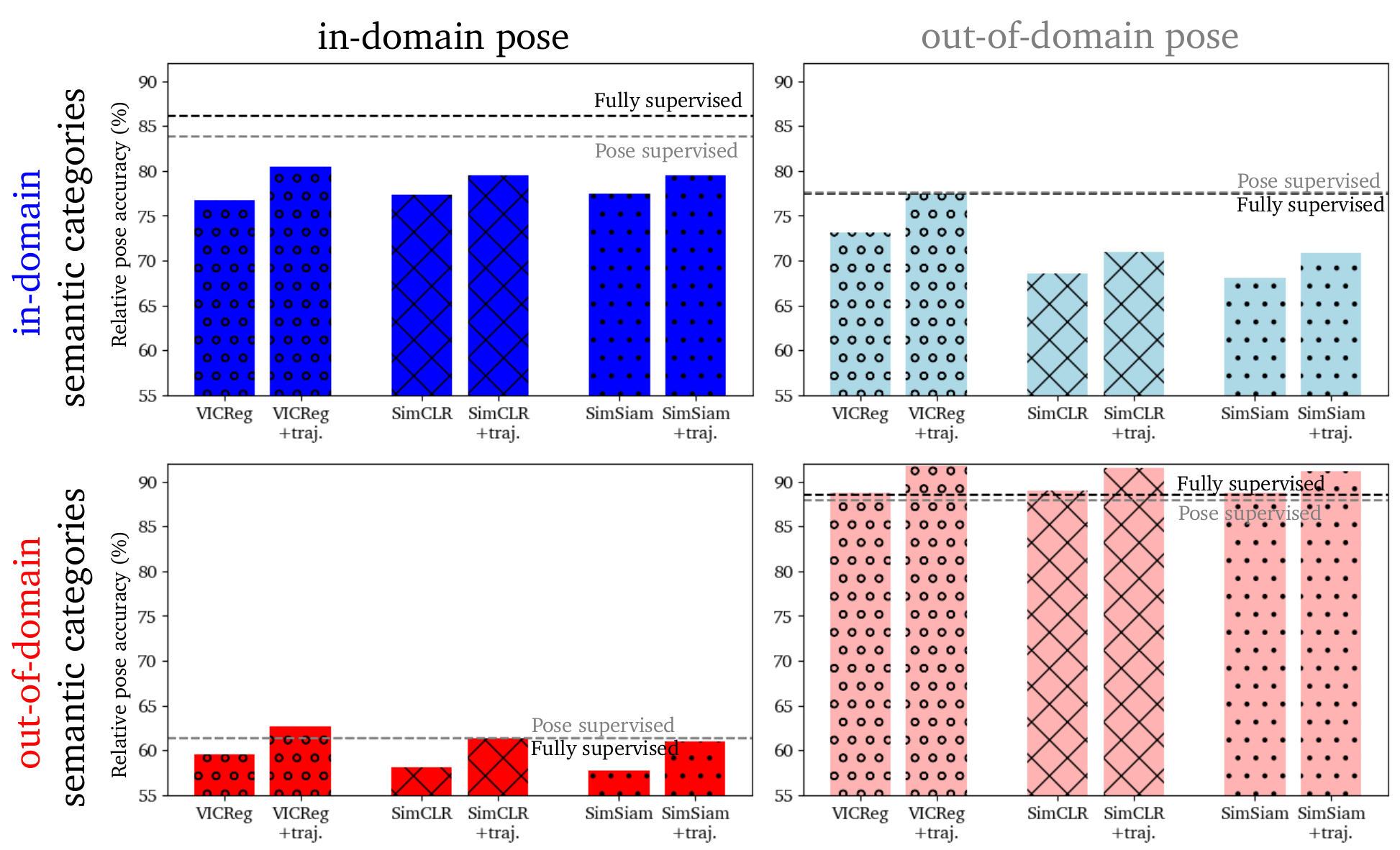}
      \vspace{-1.5em}
    \caption{  Our trajectory regularization consistently achieves higher relative pose estimation accuracy for \textcolor{blue}{in-domain}, \textcolor{red}{out-of-domain} semantic categories and \textcolor{black}{in-domain}, \textcolor{gray}{out-of-domain} poses. The bottom right figure shows the performance on real dataset \cite{shaler2017carvana}, whose high performance is due to its easier pose classification setting than simulation with Fib(50)/Fib(100) pose estiamtion. 
    Our trajectory loss $\mathcal{L}_{\text{traj}}$ leads to pose estimation gain without harming semantic classification accuracy. 
Specifically, SSL gives comparable or marginally superior results than supervised methods for out-of-domain and real data. 
Feature-layer representation $z$ is used for pose estimation.}
    \label{fig:table1}
            \vspace{-1.5em}
\end{figure}

We report the semantic classification and pose estimation performance of different methods in Fig.\ref{fig:table1}. Numerical results are in Table \ref{tab:feats} in the supplementary. For geometry-supervised and SSL methods, the same feature-layer representation $z$ is used for both geometric and semantic tasks. We aim to understand: {\bf 1)} if adding trajectory loss $\mathcal{L}_{\text{traj}}$ (Eqn.\ref{eq:traj})  helps pose estimation, and {\bf 2)} what is the gap between SSL and supervised methods.

\noindent {\bf Semantic Classification}. All methods have a similar semantic classification accuracy (85-86\%). SSL accuracies are close to the supervised upper bound. Also, adding the trajectory regularization loss $\mathcal{L}_{\text{traj}}$ leads to no accuracy loss for semantic classification, indicating that geometric representation is learned without harming semantic tasks.

\noindent {\bf In-Domain Pose Estimation}. Adding the trajectory regularization yields up to 4\% performance gain, although there is a performance gap between SSL methods and supervised methods.
Specifically, we consider two evaluation methods: absolute pose with $k$-NN and relative pose with simple probe. 
For the absolute pose estimation, adding the proposed trajectory loss leads to 4\% gain for VICReg and 2\% gain for SimCLR and SimSiam.
For the relative pose estimation, adding the proposed trajectory loss also leads to 4\% gain for VICReg and 2\% gain for SimCLR and SimSiam.
For both absolute and relative pose, SSL has a 2\%-3\% gap to geometry-supervised methods and 4\%-5\% gap to the supervised methods, which is expected as SSL takes no ground-truth pose labels.

\noindent {\bf Out-Of-Domain Pose Estimation}. Trajectory loss $\mathcal{L}_{\text{traj}}$ yields up to 4\%   gain, and SSL methods are on par or even slightly outperform supervised methods on out-of-domain pose estimation.
Specifically, for unseen poses, adding the proposed trajectory loss also leads to 3\% gain for VICReg and 3\% gain for SimCLR. SSL slightly outperforms supervised and geometry-supervised methods.
For unseen categories, adding $\mathcal{L}_{\text{traj}}$ also leads to 4\% gain for VICReg and 3\% gain for SimCLR. SSL is on par with supervised and geometry-supervised methods.
\begin{figure}[t!]
    \centering
   \includegraphics[width=1.01\columnwidth]{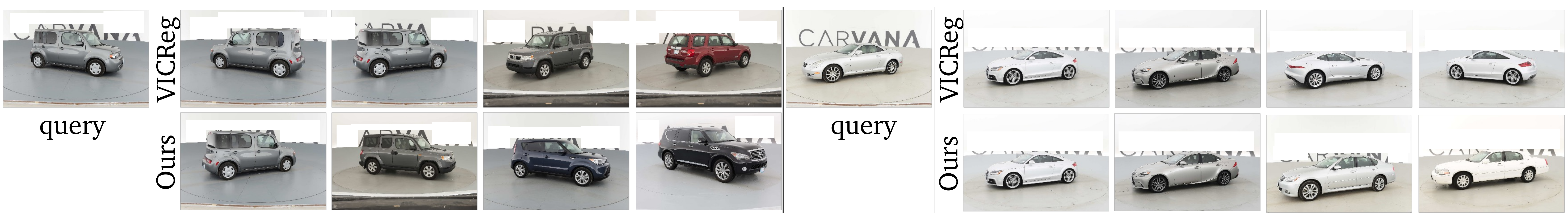}
       \vspace{-1.5em}
    \caption{
    Retrieval on a real rotating-car dataset \cite{shaler2017carvana} with 4 nearest neighbors depicted. The goal is to retrieve an image with a similar pose.
    Adding trajectory-regularization to a baseline SSL \cite{bardes2021vicreg} leads to better retrievals with similar pose and appearance, e.g. the first nearest neighbor on the left and the third nearest neighbor on the right.}
    \label{fig:retrieval}
      \vspace{-1.3em}
\end{figure}

\noindent {\bf Real Photos.} Models trained on synthetic data can directly work on real data. 
Specifically, we directly evaluate models trained on the synthetic dataset \cite{chang2015shapenet} on a real photo dataset, Carvana \cite{shaler2017carvana}, for pose estimation.
We randomly use 80\% instances as the gallery and the rest as queries.
The dataset contains $318$ car instances, each of which has 16 views, leading to $5,088$ car images in total. Adding $\mathcal{L}_{\text{traj}}$ also leads to 3\%, 3\% and 2\% gain for VICReg, SimCLR and SimSiam. SSL slightly outperforms supervised methods. Retrieval results are in Fig.\ref{fig:retrieval}.


In all, trajectory loss enhances in-domain and out-of-domain pose estimation.  SSL is on par with supervised methods on out-of-domain pose estimation.
\begin{figure*}[b]
    \centering
    \vspace{-1em}
   \includegraphics[width=1\linewidth]{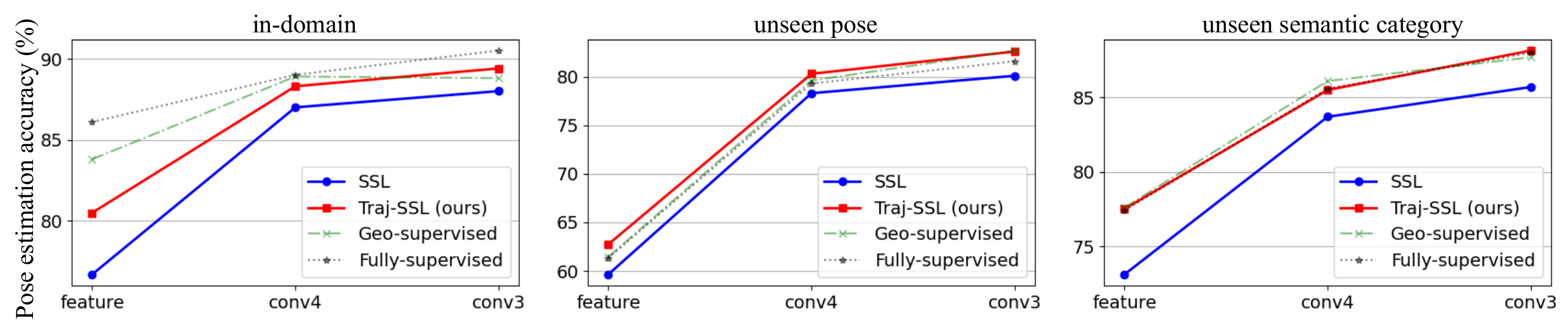}
      \vspace{-1em}
    \caption{
   Mid-layer representations improve pose estimation performance:  9\% for in-domain data,  20\% gain for out-of-domain poses and 11\% gain for out-of-domain semantic  classes. 
    SSL's gap to supervised methods is also smaller for out-of-domain data.
    }
    \label{fig:embeeding}
     \vspace{-1em}
\end{figure*}

\vspace{-1em}
\subsection{Evaluation on Mid-Layer Representations}
\label{sec:midlayer}

During training, the trajectory loss $\mathcal{L}_{\text{traj}}$ (Eqn.\ref{eq:traj}) is always constrained on feature layer $z$, as  changing the layer used for $\mathcal{L}_{\text{traj}}$ only gives $\sim 1\%$ difference (Section \ref{sec:ablation} in supplementary).
Different layers of the trained model are used as the representation for downstream geometric tasks. We report relative pose estimation performance using representations of different layers of Res18 \cite{he2016deep} and find mid-layer ``conv3'' gives the best performance (Fig.\ref{fig:embeeding}). All probes have the same parameter size. Table \ref{tab:mid} in supplementary summarizes numerical results.

\noindent {\bf In-Domain Pose Estimation}.
Using mid-layer representation ``conv3'' greatly enhances pose estimation performance over the last feature-layer. The gap is small compared with the second to the last layer ``conv-4''.
Specifically, using ``conv3'' layer as representation leads to 1\% gain over ``conv4'' and 9\% gain over `feature'' layer for VICReg with trajectory regularization. For baseline SSL and supervised methods, we also observe gain with mid-level representations.

\noindent {\bf Out-Of-Domain Pose Estimation}.  Using the mid-layer feature ``conv3''  enhances pose estimation performance on out-of-domain data, and the gap is larger for unseen poses.
Specifically, for unseen poses,  ``conv3'' layer leads to 2\% gain over ``conv4'' and 20\% gain over `feature'' layer for VICReg with trajectory regularization. 
For unseen semantic categories, ``conv3'' layer has 3\% gain over  ``conv4'' and 11\% gain over  `feature'' layer.
For baseline SSL and supervised methods, we also observe gain with mid-level representations for out-of-domain data.

\noindent {\bf Using Mid-Layer for Semantic Classification.} Empirically, we found that using ``conv3'' or ``conv4'' layer as representation for semantic classification does not make much difference (less than 1\%). 

Additional experimental results are in the supplementary.
{\bf 1)} Mid-layer representations improve the pose estimation performance but can increase the computation burden due to their high dimensionality. We show that we can compress such representation to the same dimension as the last layer with minimal performance drop (Section \ref{sec:compression} in the supplementary). 
{\bf 2)} We find that our method is robust to hyperparameters or settings including the layer to impose trajectory loss, the weight of trajectory loss and when images have non-equidistant poses. For different backbones, our method also maintains performance gain over baselines.
Refer to Section \ref{sec:ablation} in the supplementary for more details.
\begin{figure}[htb]
\vspace{-1em}
        \centering
        \includegraphics[width=0.96\linewidth]{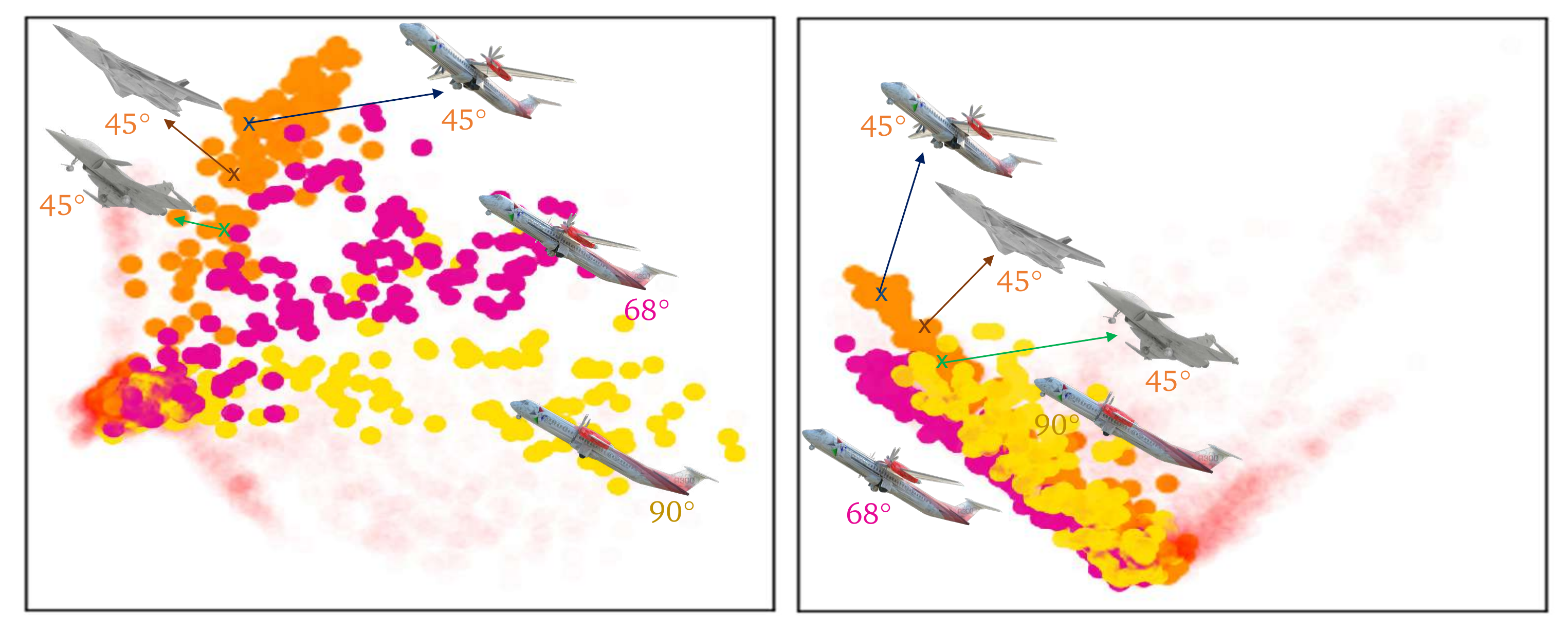} 
        \vspace{-1em}
        \caption{PCA projection of embedding of renderings of multiple airplanes with pose changes, which demonstrates the improved representation of our method ({\bf Left}) over baseline \cite{bardes2021vicreg} ({\bf Right}).In each figure, different dots refer to different airplanes with the same pose.
    We observe as airplane poses change from $(45,30^\circ)$ to $(90^\circ, 30^\circ)$, their representations form a trajectory in the feature space. While the baseline method without trajectory loss can differentiate some views, it fails to form a trajectory, which could partially contribute to worse pose estimation performance.
        }
    \label{fig:pca}
    \vspace{-1.3em}
\end{figure}

\begin{figure}[t!]
\vspace{-1.3em}
        \centering
        \includegraphics[width=\linewidth]{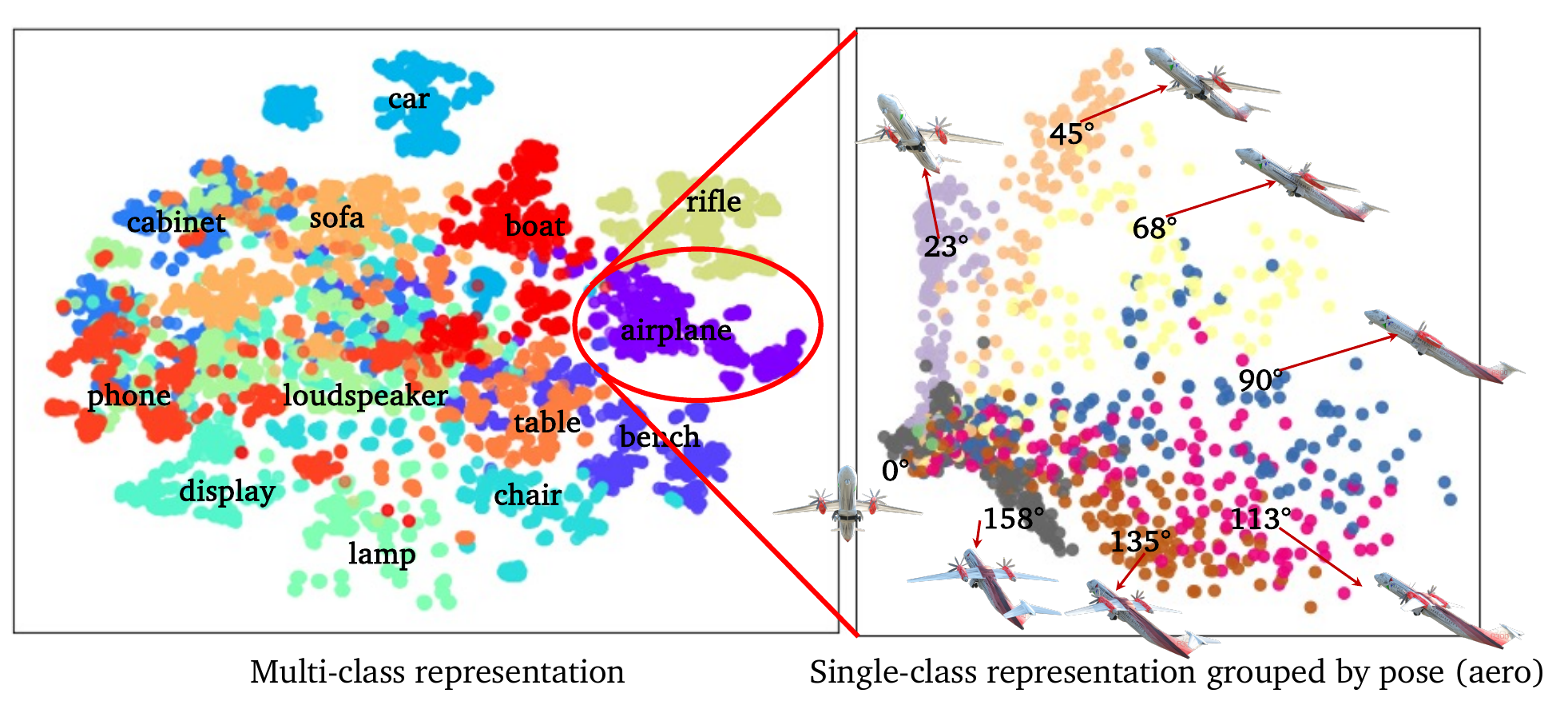} 
        \vspace{-2em}
        \caption{The joint semantic-pose embedding: Images are clustered by semantics; within each semantic cluster, images form mini-cluster by pose. 
        {\bf Left:} Representation $z$ is grouped by different semantic categories. Images with the same semantic categories form clusters. {\bf Right:} Zooming in one category, airplane, we visualize 200 instances with different poses. As the azimuth changes, their representation also forms a trajectory.}
    \label{fig:tsne}
    \vspace{-2em}
\end{figure}

\subsection{Visualizations}

{\bf Visualization of Different Poses.} 
As object poses gradually change, their representations also form a trajectory (Fig.\ref{fig:pca} and Fig. \ref{fig:pca2} in the supplementary). We visualize representations of 200 airplanes with poses ranging from $(0, 30^{\circ})$ to $ (158^{\circ}, 30^{\circ})$. These representations form a smooth trajectory. However, the baseline method without trajectory loss produces representations that can differentiate some views but may not form a coherent trajectory, which contributes to worse performance for pose estimation.

\noindent {\bf Visualization of Different Semantic Categories.} We present visualizations of the feature-layer $z$ organized by different semantic categories (Fig.\ref{fig:tsne}). We observe that images within the same semantic categories are naturally grouped together.
For a specific category, airplane, we observe that as the pose varies, the representations also cluster together, with similar poses being closer.

\noindent{\bf Summary}.
We introduce a new benchmark to evaluate geometric representations in self-supervised learning (SSL) without using semantic or pose labels during training. Our approach improves pose estimation performance by 10\%-20\% through structured and mid-level representations, with an additional 4\% gain from unsupervised trajectory regularization.

There are two major limitations: {\bf 1)} Our benchmark mainly uses synthetic data.  
{\bf 2)} While we utilize 3D pose estimation as our primary downstream task for evaluating geometric representations, the inclusion of more comprehensive geometric tasks, such as 6-DoF pose estimation or depth map prediction, could enrich the benchmark's scope and utility.
Yet, the proposed pose trajectory regularization is a general principle with the potential to benefit other geometric tasks. In conclusion, despite these limitations, our methods show potential for improving SSL's performance in geometric tasks.

\section*{Acknowledgements}
This project was supported, in part, by NSF 2215542, NSF 2313151, and Bosch gift funds to S. Yu at UC Berkeley and the University of Michigan.  The authors thank Zezhou Cheng and Quentin Garrido for helpful discussions. 

\bibliographystyle{splncs04}
\bibliography{main}
\clearpage
\setcounter{page}{1}
\appendix
\section*{Supplementary Material}

In this supplementary material, we provide details omitted in the main text including:
\begin{itemize}
\item Section \ref{sec:sdata}: Dataset statistics and comparison with similar datasets;
\item Section \ref{sec:aa}: Comparison with related methods from previous work;
\item Section \ref{sec:sresults}: Numerical results in Section \ref{sec:exp} of the main paper;
\item Section \ref{sec:compression}: Implementation and results of the mid-layer representation compression, where we compress representations with minimal performance drop;
\item Section \ref{sec:ad}: Empirical study on the similarity between mid-layer features and patch embedding;
\item Section \ref{sec:ablation}: Ablation study on the trajectory loss, non-equidistance pose of image triplets and the backbone architecture;
\item Section \ref{sec:obja}: Additional results on a large-scale dataset Objaverse \cite{deitke2023objaverse}. 
\end{itemize}
\section{More Dataset Details}
\label{sec:sdata}
\begin{figure}[htb]
    \centering
   \includegraphics[width=0.9\columnwidth]{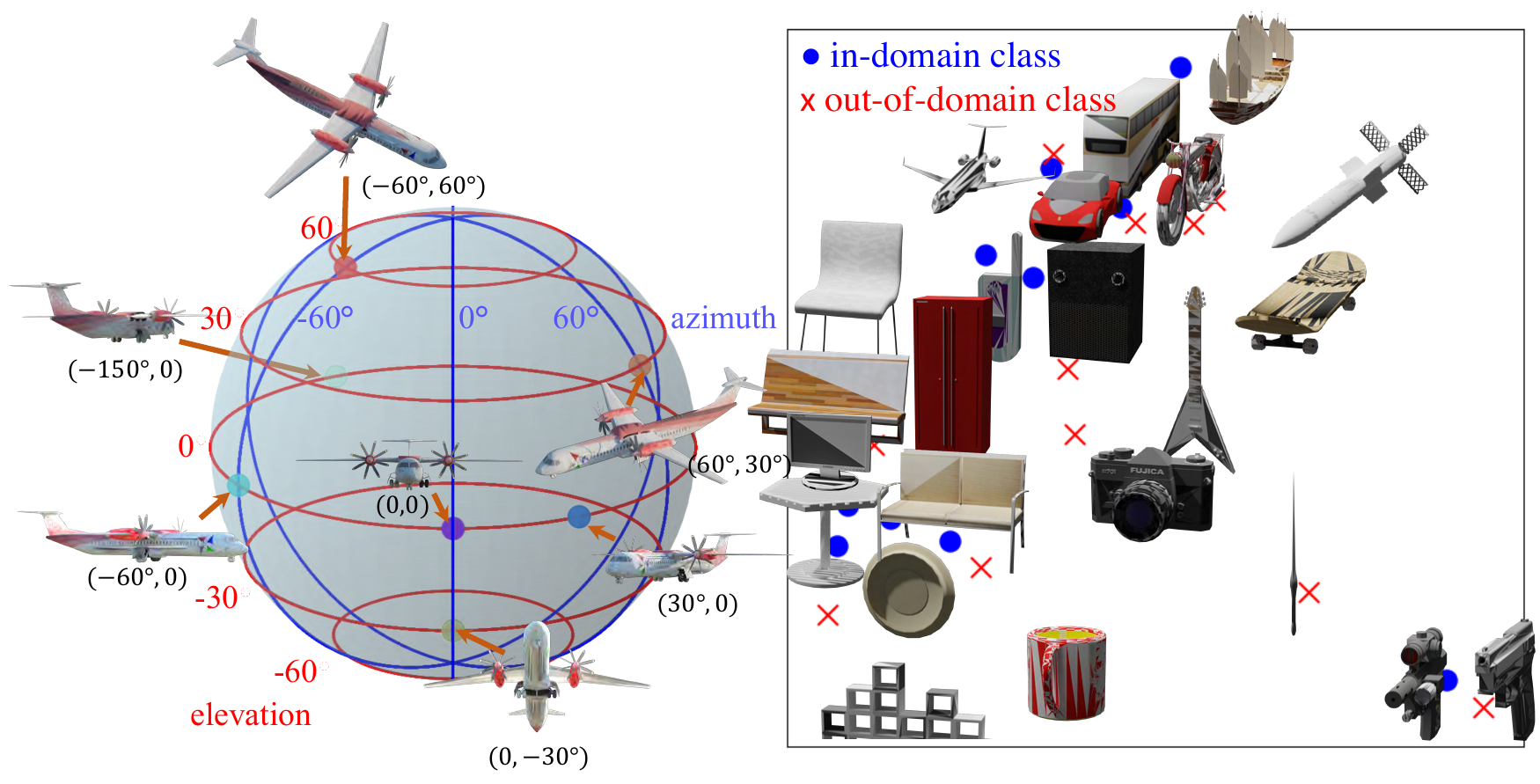}
    \caption{\small 
    {\bf Left:} We evaluate on {\it absolute} and {\it relative} pose estimation. For an object, we render images with look-at view transforms, which assume the camera is on a unit sphere with up-vector $(0, 1, 0)$ and translation $\vc{t}=(0,0,1)$. The object pose is represented as the (azimuth, elevation) of the camera angle. The figure depicts {\it absolute pose}, where we define a canonical pose for each category. For {\it relative pose}, we take two images with poses $\vc{p_1}, \vc{p_2}$ as inputs, and predict the pose difference $\vc{\Delta p} =\vc{p_2}- \vc{p_1}$.
    {\bf Right:} $(15^{\circ},15^{\circ})$ pose of \textcolor{blue}{in-domain} and \textcolor{red}{out-of-domain} semantic classes, which are plotted with PCA-projected Word2Vec \cite{church2017word2vec}.
    }
    \label{fig:relpose}
    \vspace{-1em}
\end{figure}

We divided the dataset into in-domain and out-of-domain parts, as illustrated in Table \ref{tab:dataset} and Fig.\ref{fig:relpose}. For the pose, we utilize non-overlapping Fib$(50)$ and Fib$(100)$ for in-domain and out-of-domain poses, respectively. Regarding semantic categories, we use 13 object classes (e.g., airplane, car, watercraft) as in-domain data and 11 object classes (e.g., bed, guitar, rocket) as out-of-domain data. There is no overlap between the two sets.
For each semantic category, we rendered 400 different objects, with 320 used for unsupervised training (or probe training) and 80 for testing. Our experiments are simplified by focusing on scenarios where either pose or semantics are out-of-domain, but not both.

 \begin{table}[htb]
   \begin{minipage}{0.48\textwidth}
        \centering
        \caption{
        In-domain and out-of-domain split of our benchmark dataset.
        For semantics, we use non-overlapping 13 semantic categories as in-domain and 11 as out-of-domain data. For pose, we render distinct Fib$(50)$ and Fib$(100)$ for each object as in-domain and out-of-domain poses. Only 13 in-domain categories with Fib$(50)$ poses are used for training, while others are for out-of-domain evaluation (visualizations in Fig.\ref{fig:data_vis} in the main paper).}
\small{
\resizebox{\linewidth}{!}{
\begin{tabular}{ll|l|l}
\toprule
\multicolumn{2}{l|}{\multirow{2}{*}{ \diagbox[width=12em,height=2.5em]{{\bf Pose}}{{\bf Semantic}} }}       & \textbf{\textcolor{blue}{in-domain}}   & \textcolor{red}{\textbf{out-of-domain}}     \\ 
\multicolumn{2}{l|}{}                                           & \textcolor{blue}{13 classes}  & \textcolor{red}{11 classes}        \\ \midrule
\multicolumn{1}{l}{\textcolor{black}{\textbf{in-domain}}}     & \textcolor{black}{Fib$(50)$}  & \textcolor{blue}{260k}     & \textcolor{red}{260k}         \\ 
\multicolumn{1}{l}{\textcolor{gray}{\textbf{out-of-domain}}} & \textcolor{gray}{Fib$(100)$} & \textcolor[rgb]{0.5,0.5,1}{520k}  & \textcolor[rgb]{1,0.5,0.5}{440k}   \\ 

\bottomrule
\end{tabular}
}
}
        \label{tab:dataset}
   \end{minipage}
    \hfill 
   \begin{minipage}{0.5\textwidth}
\centering
\caption{Comparison with other datasets consisting of rendered images of objects from ShapeNet \cite{chang2015shapenet}. Our dataset {\bf 1)} does not use pose labels for training and adheres to SSL geometric representation evaluation setting; {\bf 2)} enables evaluation on out-of-domain data; {\bf 2)} has complete and even pose coverage for rendered images.
}
\resizebox{\columnwidth}{!}{\begin{tabular}{l|lll}
\hline
 &{\bf Our dataset}	& {\bf 3DIEBench}&	{\bf 3DIdent}	 \\ \hline
{\bf Out-of-domain evaluation}	& Yes	&No	&No       \\\hline
{\bf Pose coverage}	&$(-\pi, \pi)$	&$(-\pi/2, \pi/2)$	&$(-\pi/2, \pi/2)$ \\
\hline
{\bf Pose sampling method} & even & uneven & uneven \\\hline
{\bf Numer of images} & 1.5M &2.5M&275k \\ \hline
\end{tabular}}
\label{tab:comp_dataset}
\end{minipage}
\end{table}

We demonstrate the configuration on the ShapeNet dataset \cite{chang2015shapenet} as an example.
There exist similar datasets derived from ShapeNet, such as 3DIEBench \cite{garrido2023self} and 3DIdent \cite{zimmermann2021contrastive}. Although such datasets are designed for or suitable for benchmarking SSL geometric representations, we still provide comparisons in Table \ref{tab:comp_dataset} given they are also derived from ShapeNet.

\section{Method Comparison}
\label{sec:aa}

 \begin{table}[b]
    \centering
        \centering
                \caption{
                Unlike supervised learning which requires labels in training, SSL uses neither semantic nor geometric labels in training and offers improved model flexibility and generalizability. Trajectory-regularized SSL further enhances geometric representations by incorporating an unsupervised geometry-trajectory-regularization loss. }
\small{
\resizebox{0.64\linewidth}{!}{
\begin{tabular}{l|c|c|c}
\toprule
{\bf Method} &	{\bf semantic label} &	{\bf pose label} &	{\bf traj. reg.} \\

 \midrule

Fully-Supervised \cite{zhang2022relpose,lin2023relpose++} &	\cmark &	\cmark	&\xmark\\
Geometry-Supervised \cite{dangovski2021equivariant,garrido2023self}	& \xmark &	\cmark	&\xmark \\
Invariant SSL (baseline) \cite{bardes2021vicreg,chen2020simple}	&\xmark &	\xmark	&\xmark\\
{\bf Trajectory-Regularized SSL (ours)}	&\xmark &	\xmark	&\cmark\\

\bottomrule
\end{tabular}
}
}
        \label{tab:method_comp}
  \end{table}

We compare with related works in Table \ref{tab:method_comp}.
For dataset, we propose a benchmark with a dataset generation/rendering configuration that {\bf 1)} adheres to the self-supervised learning (SSL) setting where neither semantic nor geometric labels are used for training;  {\bf 2)}  allows evaluation on out-of-domain data with the introduction of the relative pose.






\section{Numerical Results and Visualizations of ShapeNet}
\label{sec:sresults}

We provide detailed numerical results and visualizations omitted in the main paper due to space limit. 
Table \ref{tab:feats} provides detailed numbers of the proposed method and baselines with feature $z$ as the representation. The results are plotted in Fig.\ref{fig:table1} in the main paper.
Table \ref{tab:mid} provides detailed numbers of the proposed method and baselines with other layers as the representation. The results are plotted in Fig.\ref{fig:embeeding} in the main paper.

\begin{figure*}[t!]
    \centering
   
   \includegraphics[width=\linewidth]{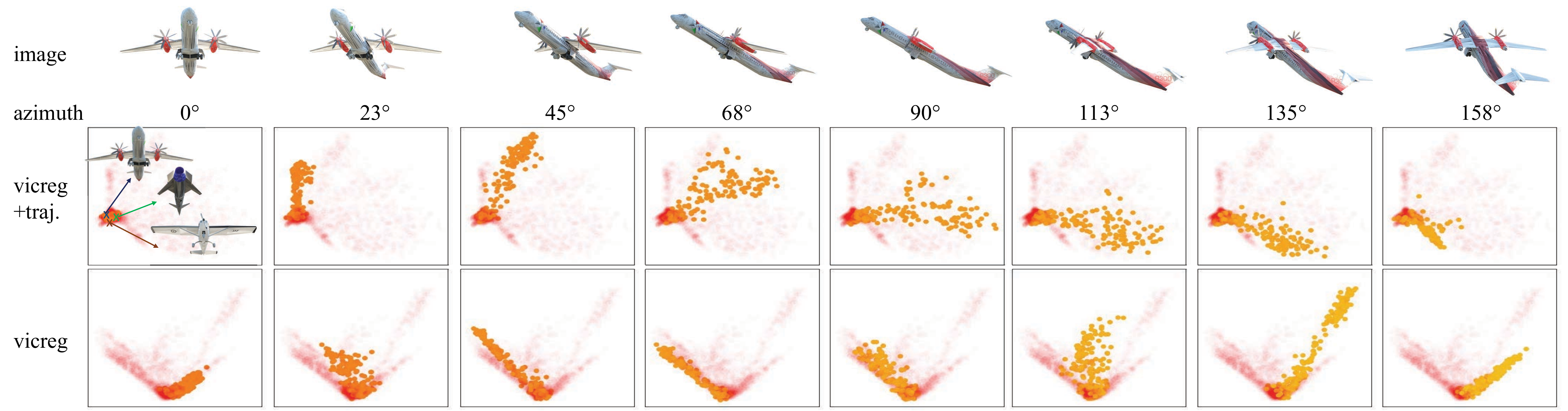}
    \caption{ PCA projection of embedding of renderings of multiple airplanes with pose changes, which demonstrates the improved representation of our method over baseline \cite{bardes2021vicreg}.
    {\bf Row 1}: Image with a pose for each column for visualizations.
    {\bf Row 2-3}:
    The embedding is the same for each row, while each column highlights multiple airplanes with the same pose.
    In each sub-figure, different dots refer to different airplanes with the same pose.
    We observe as airplane poses change from $(0,30^\circ)$ to $(158^\circ, 30^\circ)$, their representations form a trajectory in the feature space. While the baseline method without trajectory loss can differentiate some views, it fails to form a trajectory, which could partially contribute to worse pose estimation performance.
    }
    \label{fig:pca2}
\end{figure*}

Additionally, we provide a more detailed version of Fig.\ref{fig:pca2} in the main paper, where we show PCA projection of embedding of renderings of multiple airplanes with difference poses of our method and the baseline \cite{bardes2021vicreg}.

\begin{table}[htb]
    \centering
   \begin{minipage}[t]{\textwidth}
        \centering
\caption{The proposed trajectory loss $\mathcal{L}_{\text{traj}}$ leads to pose estimation gain without harming semantic classification accuracy. 
Specifically, SSL gives comparable or marginally superior results than supervised methods for out-of-domain and real data. 
Feature-layer representation $z$ is used for both semantic and pose estimation.
 }
\small{
\resizebox{\linewidth}{!}{
\begin{tabular}{@{} l | lllll|llll |ll}
\toprule

\multirow{3}{*}{{Accuracy (\%)}} & \multicolumn{5}{c|}{\textbf{In-Domain}}             & \multicolumn{4}{c|}{\textbf{Out-of-Domain Pose Est.}} &\multicolumn{2}{c}{{\bf Real Photos}} \\ \cmidrule{2-12}

                                    & \textbf{sem.} & 
                                    \textbf{abs.} & {\bf our} &\textbf{rel.} & {\bf our}& \textbf{unseen}   & {\bf our}   & \textbf{unseen}& {\bf our} &{\bf Cars } & {\bf our}  \\
                                    & {\bf cls.}&{\bf pos}&{\bf gain}&{\bf pose}&{\bf gain}&{\bf sem.}&{\bf gain}&{\bf pose}&{\bf gain}& \cite{shaler2017carvana} &{\bf gain}\\
 \midrule
\textbf{Fully-Sup. \footnote{\scriptsize We train two separate supervised models for semantic classification and pose estimation, as a supervised multi-task model yields worse results than specialized, separate models.}}                 & {86.4}   &  {92.2}    &            & {86.1}           &         & 61.3    &                 & 77.4 &   &      88.5                                          \\ \midrule

{\bf Geometry-Sup.} 
& 85.4 &  89.8        &        &  83.8        &            & 61.4      &               & {77.6}        &            &     87.9                     \\
\midrule
  \multicolumn{6}{l}{\textcolor{gray}{{\it Fully-unsupervised methods}} } \\

\textbf{VICRreg \cite{bardes2021vicreg}}                     & 85.6    &   84.3    &        & 76.7          &          & 59.6          &            & 73.1    &   &     88.7                                              \\
\textbf{VICRreg+traj.}                & 85.6   &     {\bf 87.8}&\textcolor{green}{3.5}      & {\bf 80.5}&\textcolor{green}{3.8}                    & {\bf 62.7}&\textcolor{green}{3.1}                     & {\bf 77.5}&\textcolor{green}{4.4}     &         {\bf 91.7}&\textcolor{green}{3.0}                                 \\
\addlinespace[0.4em]

\textbf{SimCLR \cite{chen2020simple}}                     &  85.9  & 84.8   &             & 77.3         &           & 58.1       &             & 68.5    &              &     89.0                                    \\
\textbf{SimCLR+traj.}                &  {\bf 86.0}  &   86.4&\textcolor{green}{1.6}                & 79.5&\textcolor{green}{2.2}                     & 61.3&\textcolor{green}{3.2}                      & 71.0&\textcolor{green}{2.5}           &       91.5\textcolor{green}&\textcolor{green}{2.5}     \\
\addlinespace[0.4em]

\textbf{SimSiam \cite{chen2021exploring}}                     &  85.4  & 84.9        &        & 77.4     &               & 57.8       &             & 68.1      &            &     88.8                                    \\
\textbf{SimSiam+traj.}                &  85.5  &   87.2&\textcolor{green}{2.3}                & 79.5&\textcolor{green}{2.1}                     & 61.0&\textcolor{green}{3.2}                      & 70.8&\textcolor{green}{2.7}           &       91.2&\textcolor{green}{2.4}     \\

\bottomrule
\end{tabular}
}
}
        \label{tab:feats}
        \end{minipage}
        \end{table}

           \begin{table}[htb]
        \centering
        \caption{Using mid-layer ``conv3'' rather than last-layer ``feature''
        for relative-pose-estimation downstream task improves accuracy: 9\% for in-domain data and 20\% for out-of-domain unseen poses.
}

\small{
\resizebox{\linewidth}{!}{
\begin{tabular}{@{} l | lll|lll|lll}
\toprule

    Rel. Pose Acc (\%)   & \multicolumn{3}{c|}{\textbf{In-Domain}} & \multicolumn{3}{c|}{\textbf{Unseen Pose}} & \multicolumn{3}{c}{\textbf{Unseen Cat.}} \\   

\textbf{Representation layer}        & conv3       & conv4       & feat       & conv3        & conv4        & feat       & conv3        & conv4        & feat       \\

\midrule
\textbf{Fully-Sup.}       & {90.5}        & 89.0       & 86.1       & 81.6         & 79.3         & 61.3       & 88.0         & 85.6         & 77.4       \\
\textbf{Geometry-Sup.} & 88.8        & 88.9        & 83.8       & 82.7         & 79.6         & 61.4       & 87.7         & 86.1         & 77.6     \\
\midrule
\textbf{VICReg \cite{bardes2021vicreg}}           & 88.0        & 87.0        & 76.7       & 80.1         & 78.3         & 59.6       & 85.7         & 83.7         & 73.1       \\
\textbf{VICReg+traj.}      &  {\bf 89.4}\textcolor{green}{($\uparrow$9)}        & 88.3        & 80.5       & {\bf 82.6}\textcolor{green}{($\uparrow$20)}         & 80.3         & 62.7       & {\bf 88.2}\textcolor{green}{($\uparrow$11)}         & 85.5         & 77.5       \\

\bottomrule
\end{tabular}
}
}
\vspace{-2em}
        \label{tab:mid}
\end{table}

\section{Compressing Mid-Layer Representations}
\label{sec:compression}






\begin{table}[b]
 \centering
 \caption{Mid-layer representations have higher pose estimation accuracies but lower efficiency due to high dimensionality. We show they can be compressed to lower dimensions with minimal performance drop for absolute pose estimation. For relative pose estimation, compressed features have a larger gap (4-5\%) but outperform representations from the feature layer.}
 
\resizebox{0.8\linewidth}{!}{ \begin{tabular}{@{} l   @{\hskip 0.03in} | @{\hskip 0.05in} l| @{\hskip 0.05in} l| @{\hskip 0.05in} l}
\toprule
{\bf embedding} &   {\bf \# dim} &  {\bf abs. pose acc. (\%)} & {\bf rel. pose acc. (\%)} \\

 \midrule

conv3&  16,384& 92.5    &87.8\\
compressed conv3    &512    &91.4 \textcolor{blue}{($\downarrow$1.1)} & 82.4 \textcolor{blue}{($\downarrow$5.4)} \\
conv4   &8,192  & 91.9 &    85.2\\
compressed conv4    &512    &90.8 \textcolor{blue}{($\downarrow$1.1)}  &    81.2 \textcolor{blue}{($\downarrow$4.0)} \\
 \midrule
feature & 512&  87.8    &   77.5\\

\bottomrule

\end{tabular}}
\label{tab:compression}
\end{table}


\noindent {\bf Motivations and Methods.} 
While mid-layer representations in networks like ResNet18 offer improved pose estimation accuracy, their large dimensions lead to inefficiencies. For instance, the ``conv3'' layer's dimension is twice that of ``conv4'' and 32 times larger than the pooled ``feature'' layer, resulting in inefficiency due to high dimensionality.
To address this, we propose compressing mid-layer representations to lower dimensions using projection heads with multi-layer perceptrons. As depicted in Fig.\ref{fig:architect} of the main paper, we denote the ``conv3'' layer representation as $\vc{z}^{\text{3}}$ and the ``conv4'' layer representation as $\vc{z}^{\text{4}}$. We then use a projection head $g_{\phi}$ to reduce the dimensionality of these representations: for ``conv3'', $\vc{y}^{\text{3}} = g_{\phi}^3(\vc{z}^{\text{3}})$; and similarly for ``conv4'', $\vc{y}^{\text{4}} = g_{\phi}^4(\vc{z}^{\text{4}})$. More details are available in the supplementary.


Then the trajectory loss $\mathcal{L}_{\text{traj}}$ (Eqn.\ref{eq:traj}) can be adapted for compressed feature $y$, e.g., when using ``conv3'' as the final representation, we can use the following trajectory  loss:
\begin{align}
\mathcal{L}_{\text{traj}}^{\text{conv3}}(\vc{y^3_L, y^3_C, y^3_R})=\mathcal{L}_{\text{traj}}( g_{\phi}^3(\vc{z_L^3}), g_{\phi}^3(\vc{z_C^3}), g_{\phi}^3(\vc{z_R^3}) )
\end{align}

\noindent{\bf Results.} For fair comparison, we make the compressed mid-layer representation $y$ has dimension of $512$, the same as the dimension of feature-layer $z$.
Our findings in Table \ref{tab:compression} demonstrate that mid-layer features can be effectively condensed 32x into smaller dimensions as ``feature''-layer with only a slight reduction in performance regarding absolute pose estimation (1\%). In the case of relative pose estimation, while there is a more noticeable difference in performance (4\%-5\%) with compressed features, they still outperform the representations derived from the feature layer.

\begin{figure}[t!]\centering
\centering
  \includegraphics[width=\columnwidth]{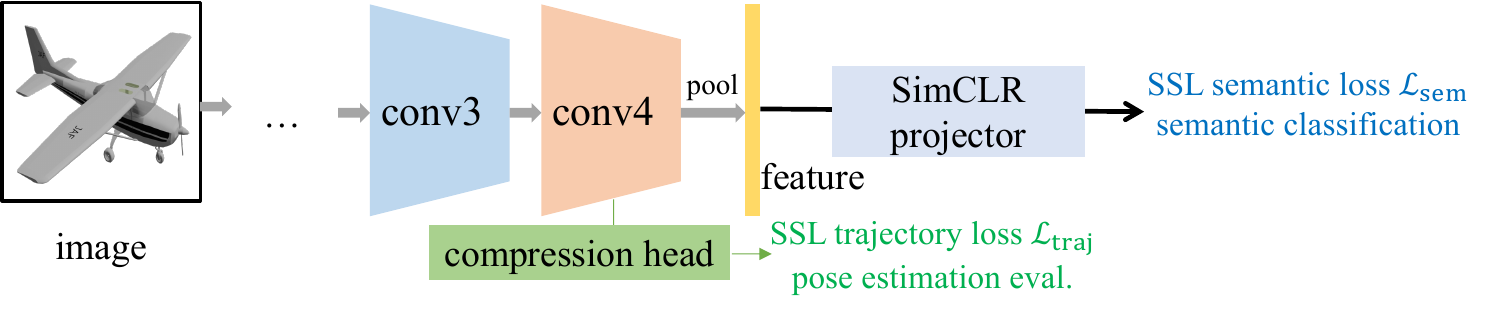}
\caption{We compress mid-layer representation from ``conv4'' layer, taking SimCLR \cite{chen2020simple} as an example. For the semantic loss, we follow SimCLR's setting and add the loss after SimCLR projector. For the pose loss,  we use an MLP-based head to compress mid-layer features and the compressed feature to classify pose. Trajectory loss is put after the compression head.
}
\label{fig:compres}
\end{figure}

{\bf Implementation Details.}
For clarity, we provide details on compressing mid-layer representations of SimCLR \cite{chen2020simple} (Fig.\ref{fig:compres}).
For the semantic loss and downstream semantic classification, we always follow the baseline setting and make no changes. We take SimCLR as an example.  
For pose estimation, we use an MLP-based head to compress mid-layer features and the compressed feature to classify pose. Trajectory is also put post-compression-head.

\section{Mid-Layer Features and Patch Embedding}
\label{sec:ad}

\begin{figure}[t!]\centering
\centering
  \includegraphics[width=0.5\columnwidth]{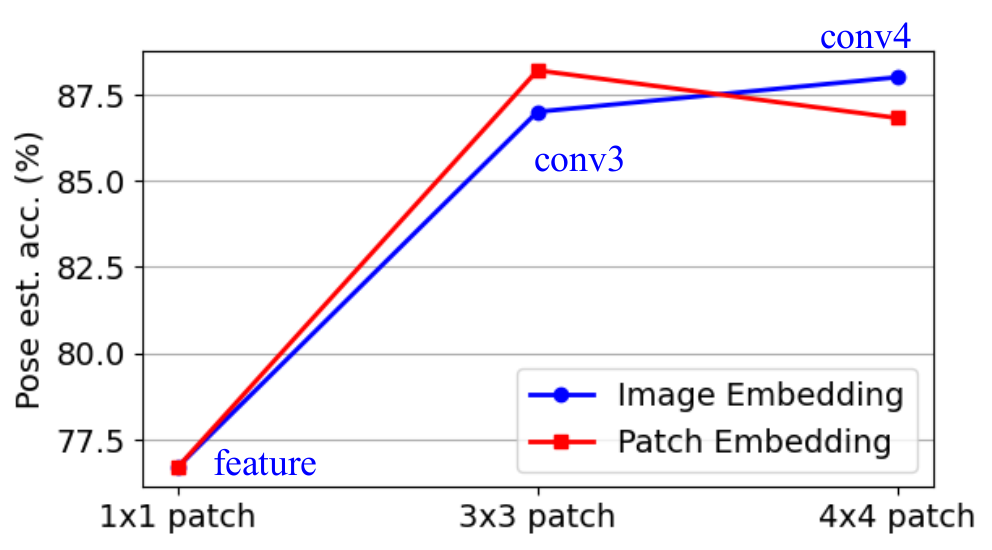}
\caption{Mid-layer representations improve SSL geometric representation quality, which could be partly attributed to the similarity to the patch embedding. Empirically, a similar trend of pose estimation accuracy gain was observed with patch embedding. The metric is relative pose estimation accuracy on in-domain data.
}
\label{fig:patche}
\end{figure}

As mentioned earlier, the improved SSL geometric representation quality by mid-layer representations could be partly attributed to the similarity to the patch embedding. Empirically, for the VICReg \cite{bardes2021vicreg} baseline, we partition the input image to $m\times m$ patches ($m=1, 3, 4$ in our experiment). As in Fig.\ref{fig:patche}, using patch embedding has a similar effect as mid-layer representation and also improves the pose estimation accuracy.

\section{Ablation Study}
\label{sec:ablation}

\begin{figure}
    \centering
     \includegraphics[width=0.8\linewidth]{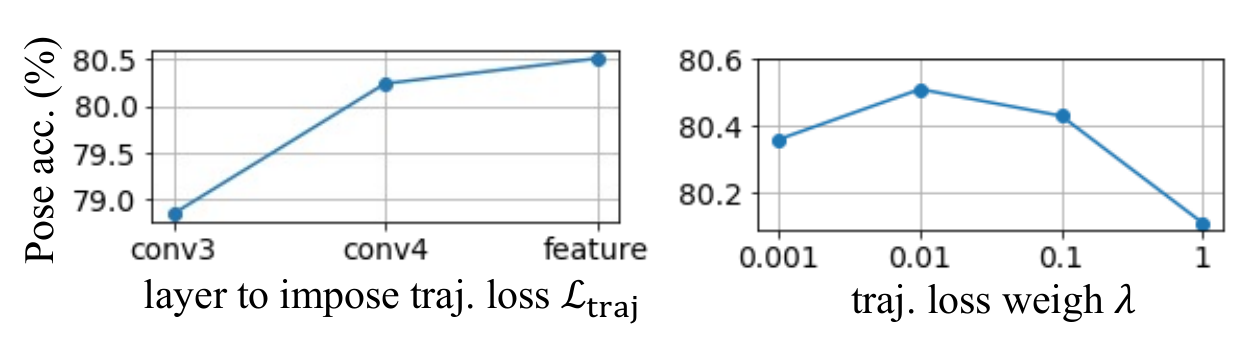}
 \caption{Hyperparameter analysis on the trajectory-regularized VICReg, which is evaluated for relative pose estimation with representation being the feature-layer $z$. {\bf Left:} While fixing the feature layer for the downstream task of pose estimation, we change different layers to impose the trajectory loss $\mathcal{L}_{\text{traj}}$. Feature-layer gives the best performance, although the difference is less than 2\%. {\bf Right:} The highest performance is achieved at trajectory loss weight $\lambda = 0.01$, though the method is not very sensitive to $\lambda$.}   
                 \label{fig:ablation}
\end{figure}

Our examination focuses on VICReg with proposed trajectory regularization, using relative pose estimation as the task and the feature layer for evaluation.  

\noindent {\bf Layer for Trajectory Loss.} In Fig.\ref{fig:ablation}{\bf U}, we vary the layer utilized for the trajectory loss $\mathcal{L}_{\text{traj}}$ during training. Note that this is different from the setting in other experiments where trajectory loss is always constrained on feature $z$ during training, and we change the layer as the representation for evaluation. The influence is $<2\%$ for different layers. 

\noindent {\bf Trajectory Loss Weight.} In Fig.\ref{fig:ablation}{\bf L}, 
the method exhibits a low sensitivity to changes in $\lambda$.


\noindent {\bf Non-Equidistant Poses}.  Our method works when the adjacent views in the trajectory loss are sampled from smooth trajectories, where the speed varies gradually.
We show this with an empirical experiment in Table \ref{tab:supp1}. Adjacent views exhibit non-equidistant poses during training: we randomly sample cubic Bézier curves with the starting pose $p_L$ and ending pose $p_R$, where the angle between $p_L, p_R$ is $(5^{\circ}, 20^{\circ})$. The middle pose $p_C$ is randomly sampled from the curve to simulate the speed variation. Non-equidistant pose trajectory regularization also gives $4\%$ gain. 

\noindent {\bf Different Backbones}.
We study if the performance gain of mid-layer representations generalizes to other network/backbone architectures.  
For VICReg \cite{bardes2021vicreg} with trajectory loss, on ResNet50 backbone we also observe a similar trend of improvement with mid-level features as the ResNet18 backbone (Table \ref{tab:supp2}).


\begin{table}[htb]
    \centering
    \begin{minipage}[t]{0.53\textwidth}
        \centering
 \centering
\caption{We render adjacent views that exhibit non-equidistant poses. Similar to equidistant poses, the trajectory loss with non-equidistant poses also gives 4\% gain for relative pose estimation. }
\resizebox{\columnwidth}{!}{\begin{tabular}{l|l}
\hline
& \textbf{Rel. pose acc(\%)} \\\hline
 {\bf VICReg} & 76.7 \\
 {\bf VICReg+equidistant traj.}&80.5\\
 {\bf VICReg+non-equidistant traj.}&80.3 \\\hline
\end{tabular}}
\label{tab:supp1}
    \end{minipage}
    \hfill 
    \begin{minipage}[t]{0.43\textwidth}
        \centering
\caption{For VICReg \cite{bardes2021vicreg} with the proposed trajectory loss, we use different backbones and also observe performance gains of relative pose estimation accuracy with mid-layer representations.}

\resizebox{\columnwidth}{!}{\begin{tabular}{l|lll}
\hline
{\bf Rel. pose acc(\%)}&feature	& conv4&	conv3 \\ \hline
{\bf Res18}	&80.5	&88.3	&89.4       \\
{\bf Res50}	&82.6	&90.1	&91.0 \\
\hline
\end{tabular}}
\label{tab:supp2}
    \end{minipage}
\end{table}

\section{Objaverse Results}
\label{sec:obja}
\begin{figure}[b]
    \begin{minipage}[t!]{0.6\textwidth}
 \centering
 \captionof{table}{Our trajectory regularization improves 1.3\% relative pose estimation accuracy; with feature layer, ours has a 3.3\% gain}
\resizebox{\linewidth}{!}{ \begin{tabular}{l|l|l|l|l}
\toprule
\textbf{Objaverse acc.}   & \multicolumn{2}{c|}{\textbf{conv4}}      & \multicolumn{2}{c}{\textbf{feat}}       \\  \midrule
\textbf{method}       & \textbf{VICReg} & \textbf{VICReg+traj.} & \textbf{VICReg} & \textbf{VICReg+traj.} \\ \midrule
\textbf{airplane}     & 86.4            & 87.0\textcolor{red}{($\uparrow$0.6)}                 & 77.9            & 81.9\textcolor{red}{($\uparrow$4.0)}                 \\
\textbf{bench}        & 90.3            & 92.1\textcolor{red}{($\uparrow$1.8)}                   & 85.0            & 88.6\textcolor{red}{($\uparrow$3.6)}                   \\
\textbf{car}          & 91.0            & 91.9\textcolor{red}{($\uparrow$0.9)}                   & 87.3            & 90.2\textcolor{red}{($\uparrow$2.9)}                   \\
\textbf{chair}        & 88.7            & 90.6\textcolor{red}{($\uparrow$1.9)}                   & 83.2            & 87.3\textcolor{red}{($\uparrow$4.1)}                   \\
\textbf{coffee table} & 88.6            & 90.0\textcolor{red}{($\uparrow$1.4)}                   & 82.0            & 84.7\textcolor{red}{($\uparrow$2.7)}                   \\
\textbf{gun}          & 81.4            & 82.8\textcolor{red}{($\uparrow$1.4)}                   & 70.6            & 73.2\textcolor{red}{($\uparrow$2.6)}                   \\
\textbf{avg}          & 87.8            & 89.1\textcolor{red}{($\uparrow$1.3)}                   & 81.0            & 84.3\textcolor{red}{($\uparrow$3.3)}       \\          
\bottomrule
\end{tabular}}
\label{stab:objaverse}

    \end{minipage}
     \hfill 
    \begin{minipage}{0.38\textwidth}
        \centering
        \includegraphics[width=1.03\linewidth]{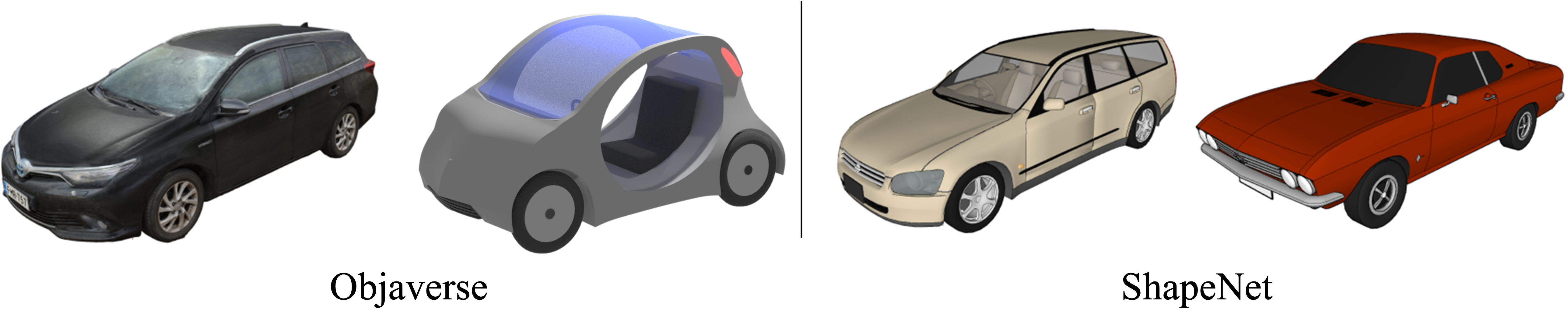} 
     \caption{Objaverse (left) has higher diversity than ShapeNet (right).}
         \label{sfig:objaverse}
    \end{minipage}

\end{figure}

We consider a 3D dataset with more diversity, Objaverse \cite{deitke2023objaverse}, with visual comparisons in Fig.\ref{sfig:objaverse}.
We carry out the experiment on a subset of Objaverse \cite{deitke2023objaverse}, and the improvement is universal on every category. The semantic categories used in this experiment: airplane, bench, car, chair, coffee table and gun. Results show that the proposed trajectory regularization is effective and using mid-layer representation helps: with conv4 layer, 
our trajectory regularization improves 1.3\% relative pose estimation accuracy; with feature layer, ours has a 3.3\% gain (Table \ref{stab:objaverse}). The full-scale Objaverse experiment with comprehensive comparison will be included in the revision. 

\end{document}